\pdfoutput=1
\documentclass[10pt,twocolumn,letterpaper]{article}

\usepackage{cvpr} 

\usepackage{graphicx}
\usepackage{amsmath}
\usepackage{amssymb}
\usepackage{booktabs}
\usepackage{float}
\usepackage{xspace}
\usepackage{url}
\usepackage[accsupp]{axessibility}
\usepackage[pagebackref,breaklinks,colorlinks]{hyperref}
\usepackage{verbatim}

\usepackage[capitalize]{cleveref}
\crefname{section}{Sec.}{Secs.}
\Crefname{section}{Section}{Sections}
\Crefname{table}{Table}{Tables}
\crefname{table}{Tab.}{Tabs.}

\def\cvprPaperID{1616} 
\def\confName{CVPR}
\def\confYear{2022}

\begin{document}

\title{Density-preserving Deep Point Cloud Compression}

\author{Yun He$^{1*}$ \quad Xinlin Ren$^{1*}$ \quad Danhang Tang$^{2}$ \quad Yinda Zhang$^{2}$ \quad Xiangyang Xue$^{1}$ \quad Yanwei Fu$^{1}$ \\
$^{1}$ Fudan University  \quad $^{2}$ Google
}

\maketitle

\newcommand{\ours}{DPCC\xspace}

\newcommand{\totalstage}{S}
\newcommand{\stage}{s}
\newcommand{\downsamplefactor}{f}
\newcommand{\feature}{\mathbf{F}}
\newcommand{\neighbors}{\mathcal{K}}

\newcommand{\point}{p}
\newcommand{\reconstructedpoint}{\hat{p}}
\newcommand{\pointcloud}{\mathcal{P}}
\newcommand{\inputpointcloud}{\pointcloud_{\stage}}
\newcommand{\downsampledpointcloud}{\pointcloud_{\stage}}
\newcommand{\reconstructedpointcloud}{\hat{\pointcloud}}

\newcommand{\dimension}{d}
\newcommand{\collapsedpointsset}{\mathcal{C}}
\newcommand{\upsampledpointsset}{\mathcal{\hat{C}}}
\newcommand{\positionembedding}{\feature^\mathcal{P}}
\newcommand{\densityembedding}{\feature^\mathcal{D}}
\newcommand{\ancestorembedding}{\feature^\mathcal{A}}
\newcommand{\fusedfeature}{\feature_{\stage}}
\newcommand{\reconstructedfeature}{\hat{\feature}}

\newcommand{\loss}{L}
\newcommand{\distortionloss}{D}
\newcommand{\rateloss}{R}
\newcommand{\rateweight}{\lambda}
\newcommand{\chamferloss}{\distortionloss_{cha}}
\newcommand{\densityloss}{\distortionloss_{den}}
\newcommand{\pointdistance}{d}
\newcommand{\pointdistanceloss}{\distortionloss_{PO}}
\newcommand{\numpointsloss}{\distortionloss_{card}}
\newcommand{\weightlatentpointcloudloss}{\mu}
\newcommand{\weightdensityloss}{\alpha}
\newcommand{\weightpointoffsetloss}{\gamma}
\newcommand{\weightnumpointsloss}{\beta}
\newcommand{\normalloss}{\distortionloss_{N}}

\newcommand{\downsamplenumber}{u}
\newcommand{\upsamplenumber}{\hat{\downsamplenumber}}
\newcommand{\maxupsamplenumber}{U}
\newcommand{\probability}{\rho}

\newcommand{\mean}[1]{\overline{#1}} 
\newcommand{\Fig}[1]{Fig~\ref{fig:#1}}
\newcommand{\Figure}[1]{Figure~\ref{fig:#1}}
\newcommand{\Table}[1]{Table~\ref{tab:#1}}
\newcommand{\eq}[1]{(\ref{eq:#1})}
\newcommand{\Eq}[1]{Eq~\ref{eq:#1}}
\newcommand{\Equation}[1]{Equation~\ref{eq:#1}}
\newcommand{\Sec}[1]{Sec~\ref{sec:#1}}
\newcommand{\Section}[1]{Section~\ref{sec:#1}}
\newcommand{\Appendix}[1]{Appendix~\ref{app:#1}}

\renewcommand{\paragraph}[1]{{\vspace{.25em}\noindent \textbf{#1.}}}

\newcommand\blfootnote[1]{%
  \begingroup
  \renewcommand\thefootnote{}\footnote{#1}%
  \addtocounter{footnote}{-1}%
  \endgroup
} 
\begin{abstract}
Local density of point clouds is crucial for representing local details, but has been overlooked by existing point cloud compression methods.
To address this, we propose a novel deep point cloud compression method that preserves local density information.
Our method works in an auto-encoder fashion: the encoder downsamples the points and learns point-wise features, while the decoder upsamples the points using these features.
Specifically, we propose to encode local geometry and density with three embeddings: density embedding, local position embedding and ancestor embedding.
During the decoding, we explicitly predict the upsampling factor for each point, and the directions and scales of the upsampled points.
To mitigate the clustered points issue in existing methods, we design a novel sub-point convolution layer, and an upsampling block with adaptive scale.
Furthermore, our method can also compress point-wise attributes, such as normal.
Extensive qualitative and quantitative results on SemanticKITTI and ShapeNet demonstrate that our method achieves the state-of-the-art rate-distortion trade-off.

\blfootnote{$^{*}$indicates equal contribution.}
\blfootnote{\,\,Yun He, Xinlin Ren and Xiangyang Xue are with the School of Computer Science, Fudan University.} 
\blfootnote{\,\,Yanwei Fu is with the School of Data Science, Fudan University.}
\end{abstract}

\begin{figure}[htb]
\centering
\includegraphics[width=\linewidth]{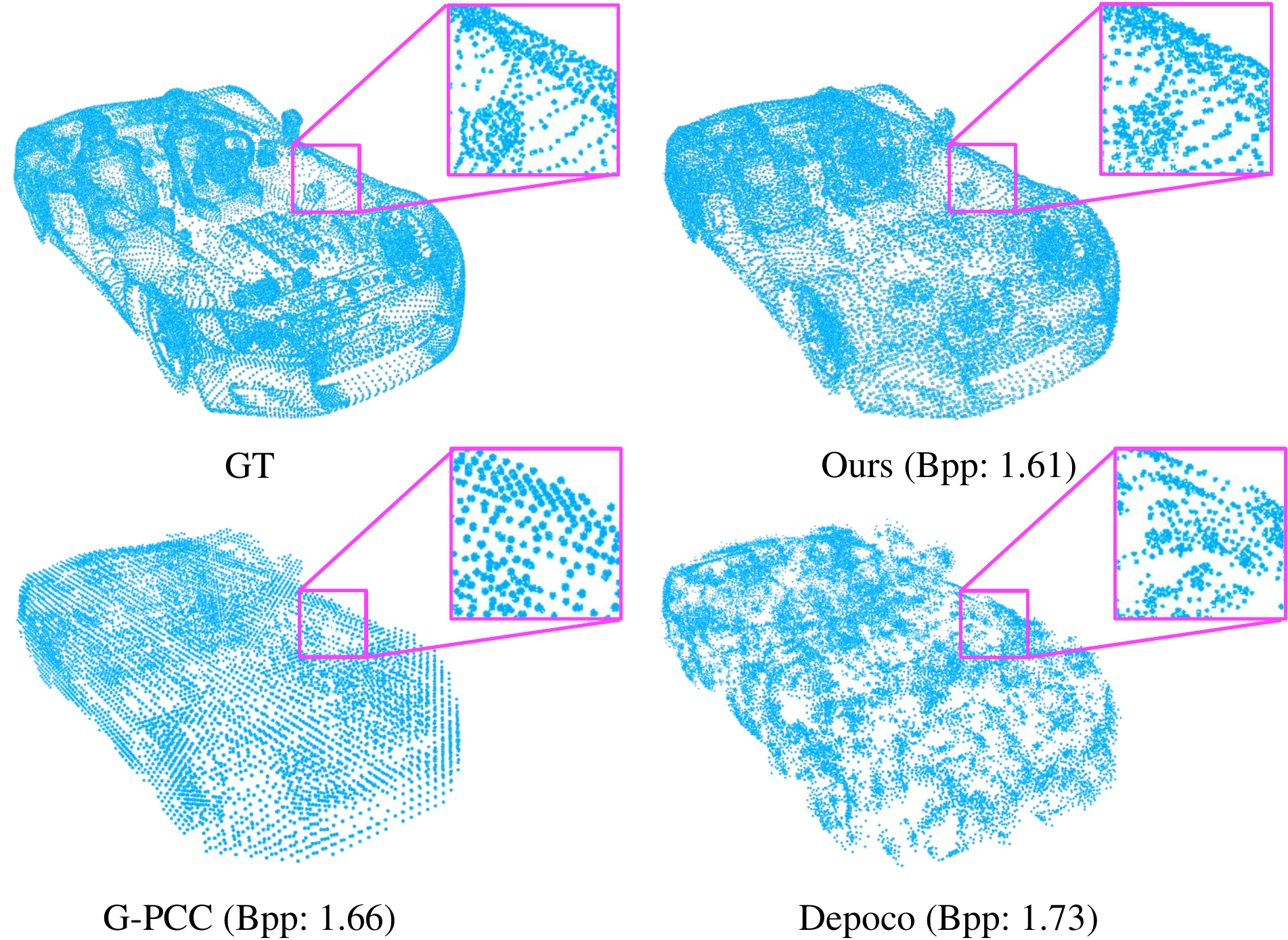}
\caption{We argue that local density is an important characteristic of the point cloud and should be preserved during compression. Existing methods that ignore the local density exhibit artifacts such as uniform distribution (G-PCC \cite{graziosi2020overview}) and clustered points (Depoco~\cite{wiesmann2021deep}), resulting in worse reconstruction, especially when the bitrate is low.}
\label{fig:teaser}
\end{figure}

\section{Introduction}

Point cloud is one of the most important and widely used 3D representation in many applications, such as autonomous driving, robotics and physics simulation \cite{guo2020deep}.
With the rapid development of 3D scanning technology, complex geometry can now be effectively captured as large point clouds with fine details.
As a consequence, point cloud compression becomes crucial for storage and transmission. Particularly, to achieve favorable compression ratio, the community has been focusing on lossy methods and pondering the key question: what properties of point clouds should be preserved, given limited bitrate budget?

Besides the global geometry, we argue that local density is  an important characteristic and should be preserved as much as possible.
Firstly, preserving density usually leads to less outliers, and thus smaller reconstruction error.
Secondly, point clouds captured in practice, \eg from LiDAR, are rarely with uniformly distributed points. Losing local density means losing important traits such as scanning resolution and occlusion. 
Thirdly, point clouds are often processed or simplified to be denser on regions of interest or with complex geometry, such as human face, hand, etc. Preserving density during compression means more budget is spent on these regions. Last but not the least, if the decompressed point cloud has significantly different density from the raw one, downstream applications such as semantic segmentation may be affected.

Mathematically, a point cloud can be considered as a set, often with different cardinality and permutation settings~\cite{bueno2021on}, which makes it difficult for image/video compression or conventional learning-based solutions that assume fixed dimensional and ordered input.
A typical strategy of existing lossy methods is to voxelize the points before compression~\cite{quach2020improved, wang2021lossy, quach2019learning, graziosi2020overview, wang2021multiscale}. While this allows leveraging conventional methods\cite{meagher1982geometric, brock2016generative}, it obviously loses the local density, and has a precision capped by the voxel size. Recent methods~\cite{yan2019deep, huang20193d} utilize PointNet~\cite{qi2017pointnet} or PointNet++~\cite{qi2017pointnet++} to ignore the cardinality and permutation with max pooling, and preserve density to some extent. 
However, the decompressed point clouds always lose local details and suffer from clustered points issue, since most of the local geometry has been discarded by max pooling.
Depoco~\cite{wiesmann2021deep} adopts KPConv \cite{thomas2019kpconv} to capture more local spatial information than pooling, but clustered points artifact still exists due to feature replication, see~\Fig{teaser}.
Alternatively, Zhao \etal \cite{zhao2021point} introduces attention mechanism to handle different cardinalities and permutations, though it is not designed for compression purpose.

In this paper, we propose a novel density-preserving deep point cloud compression method which yields superior rate-distortion trade-off to prior arts, and more importantly preserves the local density. Our method has an auto-encoder architecture, trained with an entropy encoder end-to-end. The contributions of our paper are summarized as follows. On the encoder side: 
three types of feature embeddings are designed to capture local geometry distribution and density. On the decoder side: to mitigate the clustered points issue, we propose 1) the sub-point convolution to promote feature diversity during upsampling; 2) learnable number of upsampling points, and scale for their offsets in different regions.

We conduct extensive experiments and ablation studies to justify these contributions. Additionally, we demonstrate that our method can be easily extended to jointly compress attributes such as normal.

\begin{figure*}[t!]
\centering
\includegraphics[width=\textwidth]{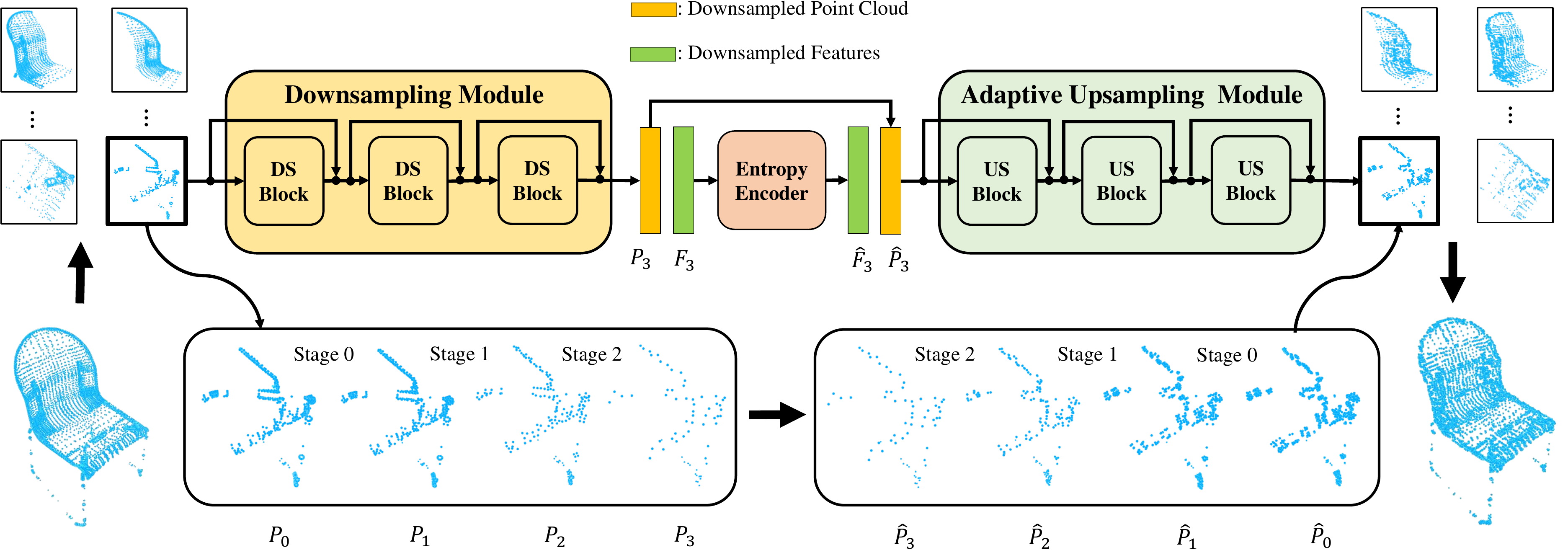}
\caption{Our pipeline first partitions the point cloud into small blocks. Each block is then downsampled three times while the local density and geometry patterns of collapsed points are encoded into features. 
At the bottleneck, downsampled features are further compressed by an entropy encoder. 
The decoder can then use the features to adaptively upsample the downsampled point cloud back to the original geometry and density.
The details of downsampling~(DS) block and upsampling~(US) block are shown in \Fig{downsample} and \Fig{upsampling_block} respectively.}
\label{fig:pipeline}
\end{figure*}

\section{Related Work}

\paragraph{Point Cloud Analysis}
Point clouds are typically unstructured, irregular and unordered, which cannot be immediately processed by conventional convolution.
To tackle this issue, many works~\cite{riegler2017octnet, maturana2015voxnet} first voxelize points and then apply 3D convolution, which however could be computationally expensive. 
Another type of approach directly operates on point clouds, hence termed point-based.
For example, 
PointNet~\cite{qi2017pointnet} and PointNet++~\cite{qi2017pointnet++} use max pooling to ignore the order of points.
DGCNN \cite{wang2019dynamic} proposes dynamic graph convolution for non-local feature aggregation.
And Point Transformer \cite{zhao2021point} introduces
a purely self-attention \cite{vaswani2017attention} based network.

\paragraph{Point Cloud Compression} 
Traditional point cloud compression algorithms~\cite{schnabel2006octree, graziosi2020overview, galligan2018google, mekuria2016design, rusu20113d, de2016compression} usually rely on octree \cite{meagher1982geometric} or KD-tree \cite{bentley1975multidimensional} structures  for storage efficiency.
Inspired by the great success of deep learning technology in point cloud analysis \cite{qi2017pointnet, qi2017pointnet++, wang2019dynamic, zhao2021point} and image compression~\cite{balle2016end, balle2018variational}, the community begins to focus on the learning based point cloud compression. 
Similarly, lossy methods can also be categorized into voxel-based~\cite{wang2021lossy, quach2019learning, quach2020improved, wang2021multiscale} and point-based\cite{yan2019deep, huang20193d, wiesmann2021deep}. While sharing the discussed pros and cons in point cloud analysis, point-based methods enable preserving local density for taking the raw 3D points as inputs.
Specifically, Yan \etal \cite{yan2019deep} integrates  PointNet~\cite{qi2017pointnet} into an auto-encoder framework, while Huang \etal \cite{huang20193d} uses PointNet++~\cite{qi2017pointnet++} instead.
Architecture wise, Wiesmann \etal \cite{wiesmann2021deep} proposes to downsample the point cloud while encoding and upsample during decoding.
Moreover, the research on deep entropy model~\cite{huang2020octsqueeze, que2021voxelcontext, biswas2020muscle} is also active, while it is nearly lossless since its loss is only from quantization.
In this paper we are focusing on the more lossy compression in favor of higher compression ratio.

\paragraph{Point Cloud Upsampling}
Point cloud upsampling aims to upsample a sparse point cloud to a dense and uniform one.
And previous methods always design various feature expansion modules to achieve it.
In particular, Yu \etal \cite{yu2018pu} replicates features and transforms them by multi-branch MLPs.
And some other methods \cite{yifan2019patch, li2019pu, li2021point} employ folding-based \cite{yang2018foldingnet} upsampling, which also duplicates features first.
Specifically, Wang \etal \cite{yifan2019patch} assigns each duplicated feature a 1D code. Li \etal \cite{li2019pu} and Li \etal \cite{li2021point} concatenate each replicated feature with a point sampled from a 2D grid.
However, the upsampled features generated from these methods could be too similar to each other due to replication, which inevitably results in clustered points.

\section{Methodology}

The proposed density-preserving deep point cloud compression framework is based on a symmetric auto-encoder architecture, where the encoder has $\totalstage$ downsampling stages indexed by $0, 1...\totalstage-1$, and the decoder also has $\totalstage$ upsampling stages indexed \textit{reversely} by $\totalstage-1, \totalstage-2...0$. For stage $\stage$ of the encoder, the input point cloud is notated as $\pointcloud_{\stage}$ and the output as $\pointcloud_{\stage+1}$. Reversely on the decoder side, the input and output of stage $\stage$ are $\reconstructedpointcloud_{\stage+1}$ and $\reconstructedpointcloud_{\stage}$ respectively, as shown in~\Fig{pipeline}. Note that to distinguish from encoding, the \textit{hat symbol} is used for reconstructed point clouds and associated features.

The input point cloud $\pointcloud_0$ is first partitioned into smaller blocks which will be compressed individually. For simplicity, we use the same notation $\pointcloud_0$ for a block.
Specifically, on the encoder side, the input $\pointcloud_\stage$ is downsampled to $\pointcloud_{\stage+1}$ by a factor $\downsamplefactor_\stage$ at each stage $\stage$, while local geometry and density are also encoded into features $\feature_{\stage+1}$.
At the bottleneck, features $\feature_{\totalstage}$ are then fed into an end-to-end trained entropy encoder for further compression.
When decompressing, we recover the downsampled point cloud $\reconstructedpointcloud_{\totalstage}$, along with the features $\reconstructedfeature_{\totalstage}$ extracted by the entropy decoder. Our upsampling module then utilizes $\reconstructedfeature_{\totalstage}$ to upsample $\reconstructedpointcloud_{\totalstage}$ back to the reconstructed point cloud $\reconstructedpointcloud_0$ stage by stage.

\subsection{Density-preserving Encoder}
\label{sec:encoder}

\paragraph{Downsampling} At each stage $\stage$ of the encoder, an input point cloud block $\pointcloud_{\stage}$ will be downsampled to $\pointcloud_{\stage+1}$ by a factor of $\downsamplefactor_\stage$ using farthest point sampling (FPS), which encourages the sampled points to have a good coverage of the point cloud $\inputpointcloud$. 
Please refer to the supplementary section for the ablation study of different sampling techniques.

\paragraph{Feature embedding}
As $\pointcloud_{\stage+1}$ itself does not preserve  the discarded points distribution of $\inputpointcloud$. 
Simply upsampling $\pointcloud_{\stage+1}$ by $1/\downsamplefactor_\stage$ will end up with a reconstruction with poor accuracy and uniform density. 
To address this, for each point $\point \in \pointcloud_{\stage+1}$, we calculate three 
different embeddings: \textit{density embedding}, \textit{local position embedding} and \textit{ancestor embedding}, to capture the geometry and density of the discarded points $\pointcloud_{\stage} - \pointcloud_{\stage+1}$ in a compact form with low entropy.

First we define the concept of a collapsed points set $\collapsedpointsset(\point)$. After the downsampled points set is decided, each discarded point is deemed to collapse into its nearest downsampled point exclusively.
Thus all the points that collapse into a downsampled point $\point$ form a collapsed points set $\collapsedpointsset(\point)$, and we term $\downsamplenumber = |\collapsedpointsset(\point)|$ as the downsampling factor of point $\point$.

The \textit{density embedding} $\densityembedding$ captures the cardinality of $\collapsedpointsset(\point)$ by mapping the downsampling factor $\downsamplenumber$ to a $\dimension$-dimensional embedding via MLPs. 
Secondly, the \textit{local position embedding} captures the distribution of $\collapsedpointsset(\point)$. Specifically, for each $\point_k \in \collapsedpointsset(\point)$, the direction and distance of the offset $\point_k-\point$ are calculated as below:

\begin{equation}
    (\frac{\point_k-\point}{||\point_k-\point||_2}, ||\point_k-\point||_2), \point \in \pointcloud_{\stage+1}, \point_k \in \collapsedpointsset (\point)
\end{equation}
where the direction (3D) and distance (scalar) are represented by this 4D vector. 
Consequently, the local point distribution centered at $\point$ can be represented by a $\downsamplenumber \times 4$ feature, which is  mapped to a higher dimensional ($\downsamplenumber \times \dimension$) space with MLPs, before attention mechanism~\cite{vaswani2017attention} is applied to aggregate them into a $\dimension$-dimensional embedding $\positionembedding$.

While the density and position embedding capture the local density and geometry at stage $\stage$, it is necessary to pass along these information from previous stages without adding much rate cost. 
To this end, we employ the point transformer layer \cite{zhao2021point} to aggregate the previous stage features of the collapsed points set $\collapsedpointsset(\point)$ into the representative sampled point $\point$, due to its simplicity and effectiveness. 
We term this $\dimension$-dimensional vector $\ancestorembedding$ as \textit{ancestor embedding}.

\begin{figure}[t!]
\centering
\includegraphics[width=\columnwidth]{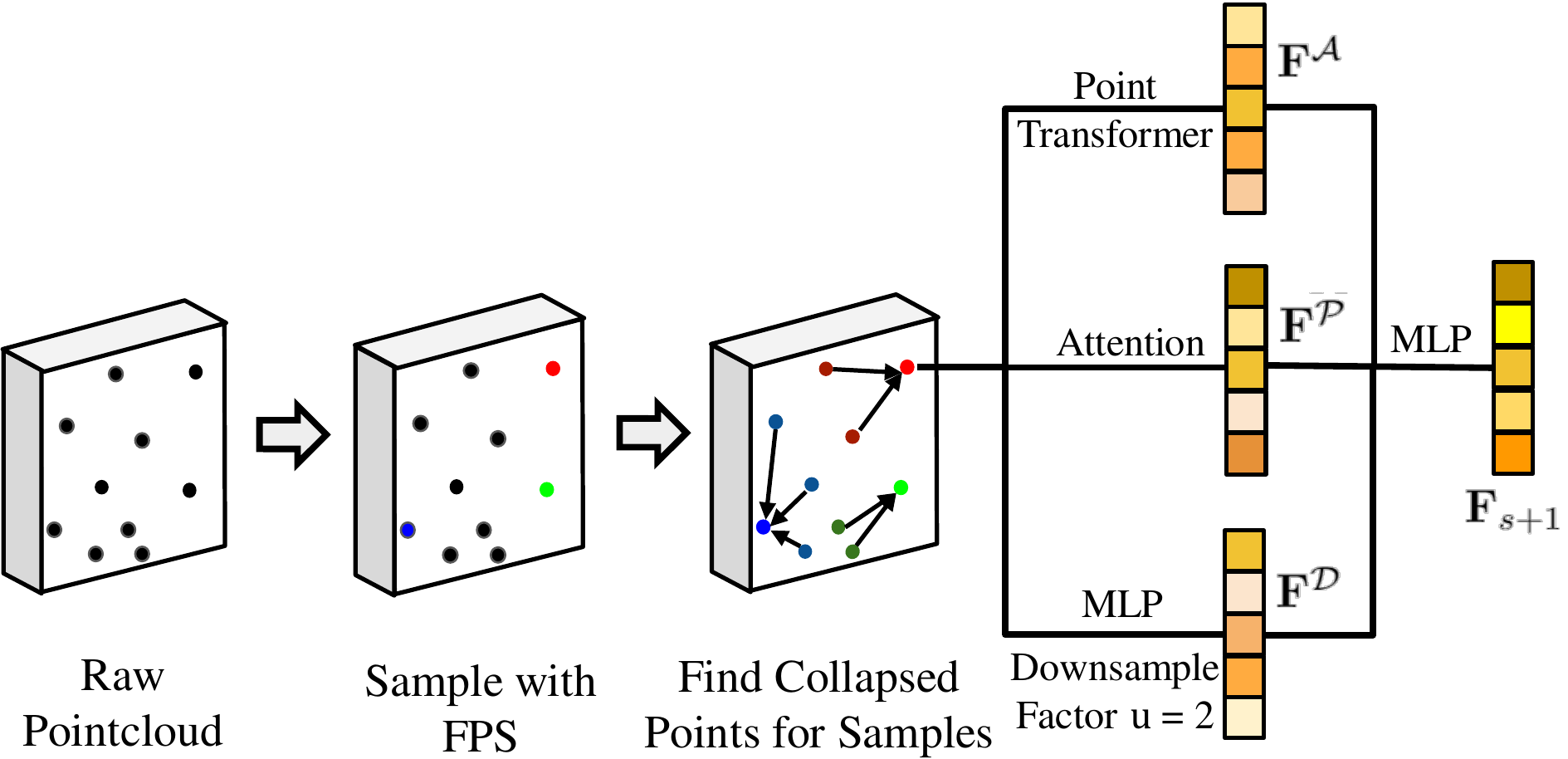}
\caption{The downsampling block: first a subset of points are chosen as samples, and then three types of embeddings are computed and fused into $\feature_{\stage+1}$.}
\label{fig:downsample}
\vspace{-10pt}
\end{figure}

At last, an MLP fuses these three embeddings $(\positionembedding, \densityembedding,\ancestorembedding)$ into a new $\dimension$-dimensional feature $\feature_{\stage+1}$ for the next stage. 
This process is illustrated in~\Fig{downsample}.

\paragraph{Entropy encoding} At the bottleneck, we have a downsampled point cloud $\pointcloud_\totalstage$ and per-point features $\feature_\totalstage$. 
For $\pointcloud_\totalstage$, we quantize it to reduce bitrate, based on \cite{begaint2020compressai}.
And $\feature_\totalstage$ are further compressed by an entropy encoder. Following recent success in deep image compression~\cite{balle2016end, balle2018variational}, we integrate an arithmetic encoder into the training process to jointly optimize the entropy of the features. This process is accompanied by a rate loss function that will be introduced later in \Sec{loss_function}.

\subsection{Density-recovering Decoder}
\label{sec:decoder}
\paragraph{Overview} During decoding, symmetrically, we have $\totalstage$ upsampling stages. 
At the bottleneck, we have the downsampled point cloud $\reconstructedpointcloud_{\totalstage}$ and decoded features $\reconstructedfeature_{\totalstage}$ extracted by the entropy decoder. 
Recall that during encoding, for each downsampled point $\point$,  $\downsamplenumber$ discarded points collapse into it. This information is not losslessly transmitted but fused into the features. 
During decoding, in order to properly upsample each point, 
we apply MLPs to predict an upsampling factor $\upsamplenumber \approx \downsamplenumber$ from the features.
Similar to the collapsed set $\collapsedpointsset(\point)$ on the encoder, we define the upsampled set of a specific point $\hat{\point} \in \reconstructedpointcloud_{\stage+1}$ as $\upsampledpointsset(\hat{\point})$.

In addition to $\upsampledpointsset(\hat{\point})$, the feature of each upsampled point is also predicted. Therefore the output of each point $\hat{\point}$ at upsampling stage $\stage$ is:
\begin{equation}
(\underset{\maxupsamplenumber\times 3}{\upsampledpointsset(\reconstructedpoint)}, \underset{\maxupsamplenumber\times \dimension}{\reconstructedfeature(\reconstructedpoint)}, \upsamplenumber), \reconstructedpoint \in \reconstructedpointcloud_{\stage+1}, \upsamplenumber \leq \maxupsamplenumber
\end{equation}
where $\upsampledpointsset(\reconstructedpoint)$ and  $\reconstructedfeature(\reconstructedpoint)$ here have $\maxupsamplenumber$ items, but only the first $\upsamplenumber$ points and features will be chosen as the final outputs. The union of all chosen points is the upsampled point cloud $\reconstructedpointcloud_{\stage}$ for the next stage, and same goes for $\reconstructedfeature_{\stage}$.

\begin{figure}[t!]
\centering
\includegraphics[width=0.9\columnwidth]{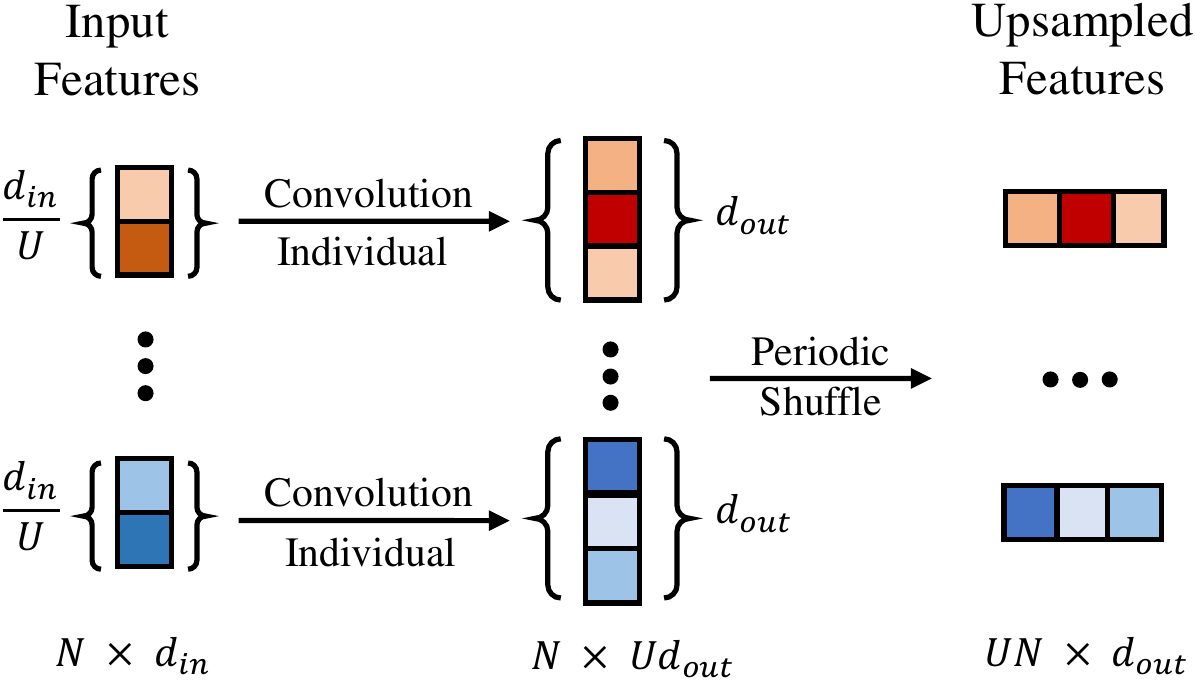}
\caption{The illustration of sub-point convolution.}
\label{fig:sub-point_convolution}
\vspace{-10pt}
\end{figure}

\paragraph{Sub-point convolution}
At upsampling stage $\stage$, guided by the features $\reconstructedfeature_{\stage+1}$, we aim to upsample each point $\reconstructedpoint \in \reconstructedpointcloud_{\stage+1}$ by the predicted upsampling factor $\upsamplenumber$. Additionally, $\reconstructedfeature_{\stage+1}$ also need to be expanded to $\upsamplenumber$ features ${\reconstructedfeature_{\stage}}$ for the next stage. To achieve so, prior upsampling methods either use multi-branch MLPs for feature expansion~\cite{yu2018pu, wiesmann2021deep} or apply folding-based \cite{yang2018foldingnet} upsampling modules~\cite{yifan2019patch, li2019pu, li2021point}. 
Despite efforts of regularization and refinement, they still suffer from the aforementioned \textit{clustered points} artifact due to feature replication.
To address this, we propose a novel and efficient operator \textit{sub-point convolution}~(\Fig{sub-point_convolution}), inspired by the sub-pixel convolution~\cite{shi2016real}.

Specifically, given the input $N \times \dimension_{in}$ features $\reconstructedfeature_{\stage+1}$, we first divide them into $\maxupsamplenumber$ groups along the channel dimension, such that each group has $\dimension_{in}/\maxupsamplenumber$ channels.
A convolution layer per group is applied to expand the features to a space with dimension $N \times \maxupsamplenumber\dimension_{out}$.

At last, we use periodic shuffle to reshape the upsampled features to $\maxupsamplenumber N \times \dimension_{out}$.
Compared with prior methods \cite{yu2018pu, yifan2019patch, li2019pu, li2021point, wiesmann2021deep}, sub-point convolution has the following advantages: 1) the clustered points issue is mitigated by preventing feature replication;
2) convolution is applied to each group with lower dimension, which significantly reduces the parameters and computations.

\begin{figure}[htbp]
\centering
\includegraphics[width=\columnwidth]{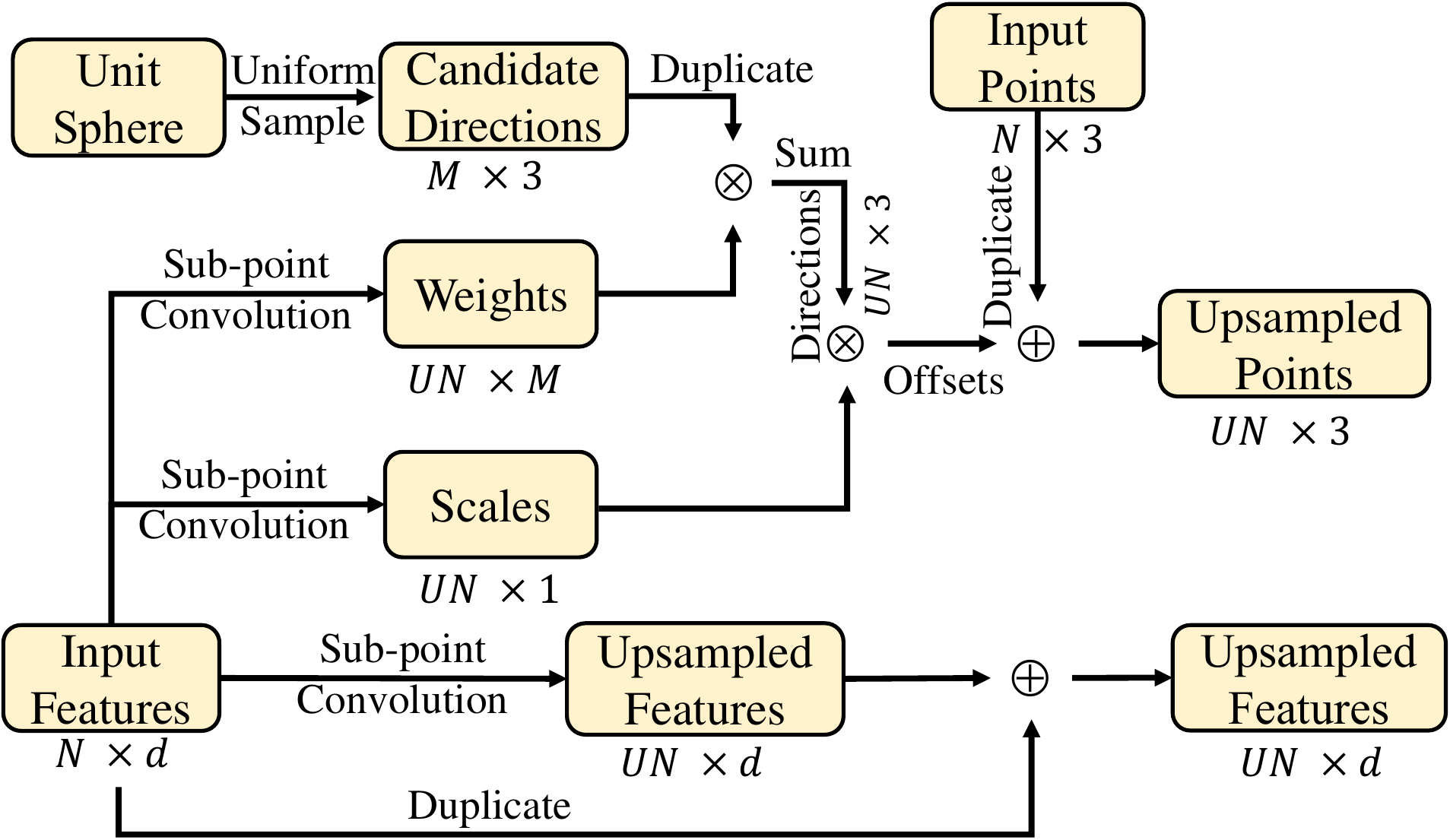}
\caption{The scale-adaptive upsampling block, includes both point upsampling and feature upsampling.}
\label{fig:upsampling_block}
\end{figure}

\paragraph{Upsampling block with adaptive scale}
Based on the sub-point convolution, we build our upsampling block for points and associated features, as depicted in \Fig{upsampling_block}.
Centering at each point $\reconstructedpoint \in \reconstructedpointcloud_{\stage+1}$,  offsets of upsampled points are predicted. Since both downsampling and upsampling happen in local regions, the scales of predicted offsets need to be constrained.
To this end, folding-based methods \cite{yifan2019patch, li2019pu, li2021point} use predefined small grid sizes. While Wiesmann \etal \cite{wiesmann2021deep} constrains predicted offsets to [-1,1], and then scales them with a predefined factor.
However, this scaling factor may vary significantly across different regions and different point clouds. Hence we design a new upsampling module with learnable scales.

In particular, a pool of $M$ vectors is first sampled from a unit sphere and kept fixed as candidate directions for both training and inference. During upsampling, weights of these candidates are predicted such that the weighted sum result is the most probable direction. Some scaling factors, or magnitudes are also predicted from the input features $\reconstructedfeature_{\stage+1}$ to have the offsets and thus upsampled points.

The feature expansion is performed by sub-point convolution within a residual block \cite{he2016deep}. 
Once we obtain the final points and features, a refinement layer is added to finetune the upsampled points and features.
It is essentially an upsampling block with upsampling factor $\upsamplenumber=1$.

\subsection{Loss Function} \label{sec:loss_function}
We employ the standard rate-distortion loss function during training for better trade-off.
\begin{equation}
    \loss = \distortionloss + \rateweight \rateloss,
\end{equation}
where $\distortionloss$ penalizes distortion and $\rateloss$ penalizes bitrate.

\paragraph{Distortion loss}
For distortion (reconstuction error), we utilize the symmetric point-to-point Chamfer Distance \cite{huang2020octsqueeze} to measure the difference between the reconstructed point cloud $\reconstructedpointcloud_\stage$ and ground truth $\pointcloud_\stage$. Since the decoder has $\totalstage$ stages, to avoid error accumulation, we compute the distortion loss at each stage and aggregate them as $\chamferloss$.

A density term is also designed to encourage recovering local density.  At stage $\stage$ of the decoder, a point $\reconstructedpoint$ is upsampled to a new chosen points set $\upsampledpointsset(\reconstructedpoint)$~(see \Sec{decoder}).
We then find its nearest counter point $\point$ on the encoder side, which is collapsed from a set $\collapsedpointsset(\point)$~(see \Sec{encoder}).
Hence we can define the density loss $\densityloss$ as:
\begin{equation}
    \densityloss = \sum_{s=0}^{\totalstage-1}
    \sum_{\reconstructedpoint \in \reconstructedpointcloud_{\stage+1}}
    \frac{
    \left||\collapsedpointsset(\point)| - |\upsampledpointsset(\reconstructedpoint)| \right| + \gamma \left|\mean{\collapsedpointsset(\point)} - \mean{\upsampledpointsset(\reconstructedpoint)}\right|}{|\reconstructedpointcloud_{\stage+1}|}
\end{equation}
where the first term of numerator calculates the cardinality difference between the two sets, the second calculates the difference between the mean distances of all points in sets to center points $\point$ or $\hat{\point}$, and $\gamma$ is the weight.

To further facilitate the density estimation, for each stage $\stage$, we utilize another loss to measure the cardinality difference of ground truth $\pointcloud_{\stage}$ and reconstructed point cloud $\reconstructedpointcloud_{\stage}$:
\begin{equation}
    \numpointsloss = \sum_{s=0}^{\totalstage-1} \left||\pointcloud_{\stage}| - |\reconstructedpointcloud_{\stage}|\right|
\end{equation}

Finally, the overall distortion loss is as follows:
\begin{equation}
    \distortionloss = \chamferloss +\weightdensityloss \densityloss + \weightnumpointsloss \numpointsloss
\end{equation}
where $\weightdensityloss$ and $\weightnumpointsloss$ are the weights of respective terms.

\paragraph{Rate loss}
Since entropy encoding is non-differentiable, a differentiable proxy is applied during training. Following~\cite{balle2016end, balle2018variational}, we replace the quantization step with an additive uniform noise, and estimate the number of bits as the rate loss $\rateloss$.
During inference, features are properly quantized and compressed by a range encoder.

\subsection{Attribute Compression}
Our framework can also compress point cloud attributes such as color, normal, etc. As an example, we incorporate normal compression into our framework. To avoid extra cost of bitrate, we fix the same network architecture and hyperparameters. The only difference is the input/output dimension has changed from 3D to 6D (position+normal). To facilitate this, we employ a simple L2 loss to minimize the normal reconstruction error.

\section{Evaluation}

\begin{figure*}[t!]
\centering
\includegraphics[width=0.93\textwidth]{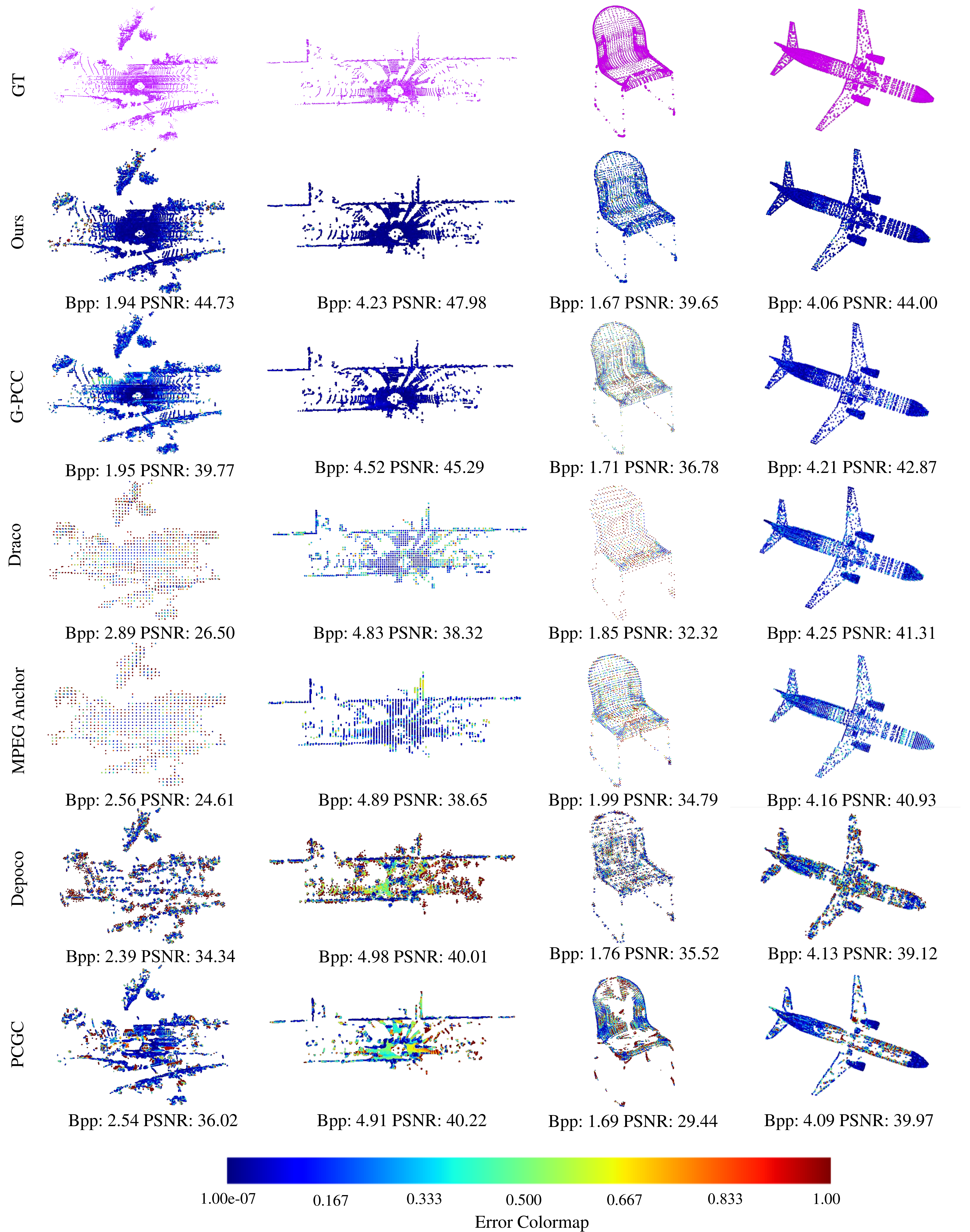}
\caption{Qualitative results on SemanticKITTI (the first two columns) and ShapeNet (the last two columns). 
From top to bottom: Ground Truth, Ours, G-PCC~\cite{graziosi2020overview}, Draco~\cite{galligan2018google}, MPEG Anchor \cite{mekuria2016design}, Depeco~\cite{wiesmann2021deep} and PCGC~\cite{wang2021lossy}.
We utilize the distance between each point in decompressed point clouds and its nearest neighbor in ground truth as the error.
And the Bpp and PSNR metrics are averaged by each block of the full point clouds.
It is obvious that our method successfully achieves both the most accurate geometry and lowest bitrates.}
\label{fig:quality_result}
\end{figure*}

In this section, we evaluate our method by comparing to state-of-the-art methods on compression rate, reconstruction accuracy and local density recovering.
We then provide ablation studies to justify the design choices.
Lastly, we demonstrate that additional attributes like normal can be also compressed. Please refer to the supplementary section for implementation details and parameter settings.

\begin{figure*}[t!]
\centering
\includegraphics[trim={0cm, .4cm, 0cm 0cm}, clip, width=\textwidth]{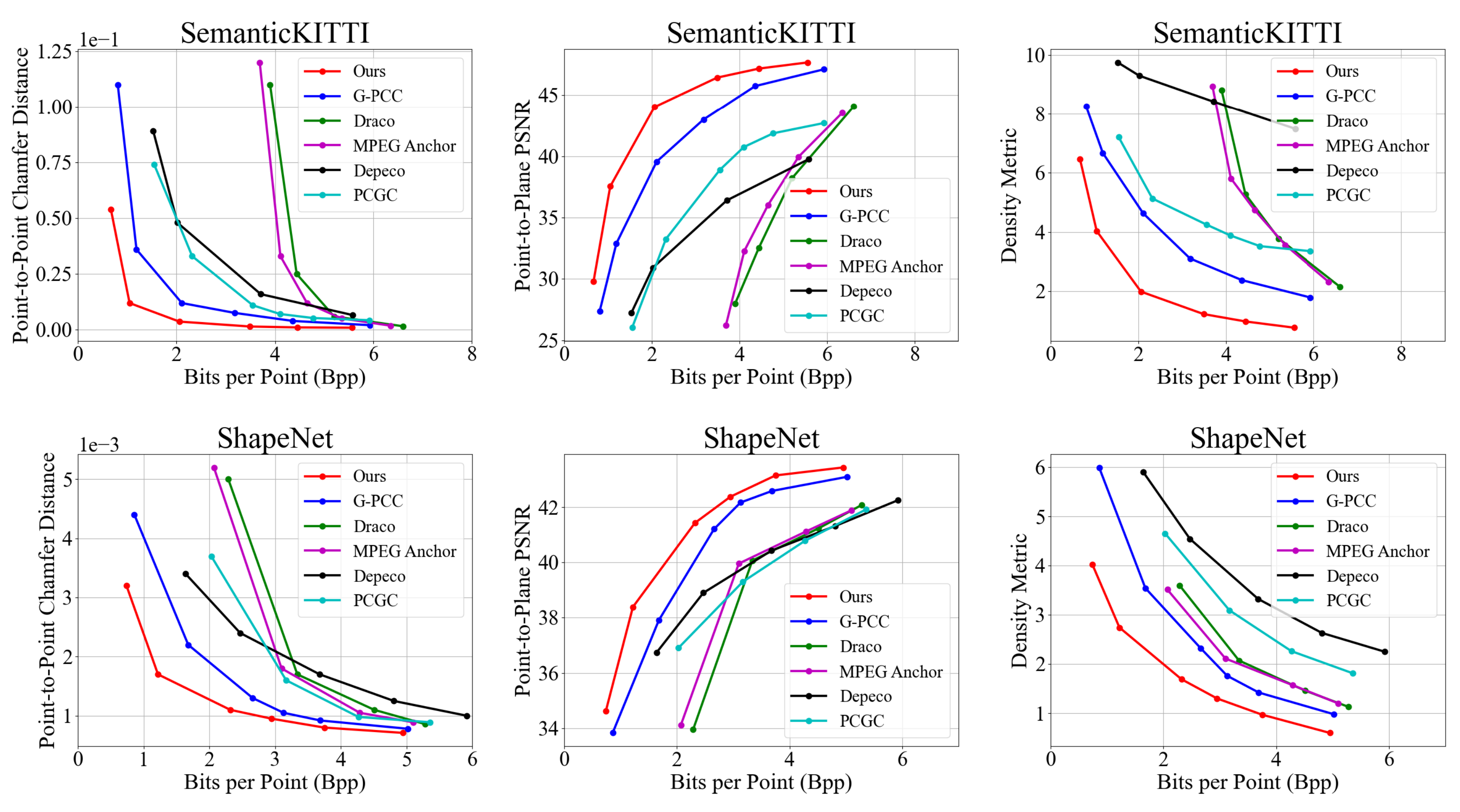}
\caption{Quantitative results on SemanticKITTI (the first row) and ShapeNet (the second row). 
Our method consistently achieves more accurate geometry and recovering density across the full range of bitrates.}
\label{fig:location_quantitive}
\vspace{-10pt}
\end{figure*}

\subsection{Experiment Setup}   
\paragraph{Datasets} 
We conduct our main experiments on SemanticKITTI  \cite{behley2019semantickitti} and ShapeNet \cite{chang2015shapenet}.
For SemanticKITTI, we utilize the training/testing split in \cite{wiesmann2021deep}.
For ShapeNet, we obtain the watertight manifold surfaces by \cite{huang2018robust}
and sample points from these meshes based on \cite{hermosilla2018monte, peters1997simplest}.  
All point clouds are first normalized to 100$m^3$ cubes and divided into non-overlapping blocks of $12m^3$ and $22m^3$ for SemanticKITTI and ShapeNet respectively, while each block is further normalized to [-1, 1]. 
For downstream surface reconstruction task, we use the RenderPeople \cite{renderpeople} dataset.

\paragraph{Baselines} 
We compare to both state-of-the-art non-learning based methods: G-PCC~\cite{graziosi2020overview}, Google Draco~\cite{galligan2018google}, MPEG Anchor \cite{mekuria2016design}; and learning-based methods: Depeco~\cite{wiesmann2021deep}, PCGC~\cite{wang2021lossy}.
Note that all learning-based methods have been retrained on the same datasets as our method.

\paragraph{Evaluation metrics}
Following \cite{biswas2020muscle,huang2020octsqueeze}, we adopt the symmetric point-to-point Chamfer Distance (CD) and point-to-plane PSNR for geometry accuracy and Bits per Point~(Bpp) for compression rate.
Moreover, we design a new metric to measure the local density differences.
And all these metrics are evaluated on each block.
Specifically, for each point $\point$, we notate its neighbor points within radius $r=0.15$ as $\neighbors(\point)$.
Since the cardinalities of ground truth $\pointcloud_0$ and reconstructed point cloud $\reconstructedpointcloud_0$ are not necessarily the same, we define a symmetric density metric $DM$ as:
\begin{equation}
\begin{aligned}
    DM(\pointcloud_0, \reconstructedpointcloud_0) &=
    \frac{1}{|\pointcloud_0|} \sum_{p \in \pointcloud_0} \delta(p, \hat{p}) + \frac{1}{|\reconstructedpointcloud_0|} \sum_{\hat{p} \in \reconstructedpointcloud_0} \delta(\hat{p},p),\\
    \textrm{where}~ \delta(a, b) &= \frac{\left||\neighbors(a)| -|\neighbors(b)|\right|}{|\neighbors(a)|} + \mu \frac{\left|\mean{\neighbors(a)} - \mean{\neighbors(b)}\right|}{\mean{\neighbors(a)}}
\end{aligned}
\end{equation}
where $b$ is the nearest counter point of $a$, $\mu$ is the weight,
$|\neighbors(a)|$ denotes the cardinality of $\neighbors(a)$ and $\mean{\neighbors(a)}$ denotes the mean distance of all points in $\neighbors(a)$ to $a$.

\subsection{Comparison with SOTA}
We first compare our method with SOTA on the rate-distortion trade-off.
In \Fig{location_quantitive}, we show the per-block Chamfer Distance, PSNR and density metric of all methods against Bits per Point (Bpp).
Our method yields more accurate reconstruction consistently across the full spectrum of Bpp on both SemanticKITTI and ShapeNet datasets.
Note the differences are more evident under the density metric.

\Fig{quality_result} shows qualitative results at various bitrates.
Draco \cite{galligan2018google} and MPEG Anchor \cite{mekuria2016design} typically need a high Bpp (\eg \textgreater 4) to achieve a satisfactory reconstruction. Plus they perform poorly at low bitrates due to quantization.
Depoco \cite{wiesmann2021deep} often generates clustered points caused by feature replication. 
PCGC \cite{wang2021lossy} tends to miss a continuous chunk of points, because it regards decompression as a binary classification process (occupied or not), which has extremely imbalanced data due to the intrinsic sparsity of point clouds. 
Besides, it also significantly alters the density.
Although G-PCC \cite{graziosi2020overview} recovers the overall geometry successfully, due to voxelization, it loses local details. 
Our method achieves the highest compression performance in terms of both geometry and local density while spending the lowest bitrates.

\begin{table}[htbp]
\centering
\resizebox{\linewidth}{!}{
\begin{tabular}{llll}
\toprule
Methods & Enc. time~(ms)  & Dec. time~(ms) & Size~(MB)\\
\midrule
G-PCC~\cite{graziosi2020overview} & 180/165 & 163/152 & 3.49 \\
Draco~\cite{galligan2018google} & 147/153 & 147/153 & 2.49 \\

MPEG Anchor~\cite{mekuria2016design} & 151/142 & 136/130 & 27.8 \\
Depoco~\cite{wiesmann2021deep} & \textbf{32}/126 & \textbf{2}/\textbf{2} & \textbf{0.38} \\
PCGC~\cite{wang2021lossy} & 130/96 & 24/19 & 7.73 \\
\midrule
Ours & 80/\textbf{81} & 24/31 & 0.44\\
\bottomrule
\end{tabular}}
\vspace{-5pt}
 \caption{The average per-block encoding time, decoding time and model size of different methods on SemanticKITTI/ShapeNet, using a TITAN X GPU.
 }
\label{tab:complexity}
\vspace{-10pt}
\end{table}

\paragraph{Complexity analysis}
\Table{complexity} shows the per-block latency and memory footprint of different methods. For G-PCC \cite{graziosi2020overview}, Draco \cite{galligan2018google} and MPEG Anchor \cite{mekuria2016design}, we use the sizes of their executable files.
For Depoco \cite{wiesmann2021deep} and PCGC \cite{wang2021lossy}, we use their checkpoint sizes. 
Our model is competitive in computational efficiency, only second to Depoco~\cite{wiesmann2021deep} but achieves a better rate-distortion trade-off.

\subsection{Ablation Study} 
For fair comparison, we conduct all the ablation experiments on SemanticKITTI while fixing the Bpp at 2.1.

\paragraph{Effectiveness of each component}
We build a baseline model consisting of a point transformer encoder \cite{zhao2021point}, entropy encoder and multi-branch MLPs decoder\cite{yu2018pu}. The proposed components, including dynamic upsampling factor $\upsamplenumber$, local position embedding $\positionembedding$, density embedding $\densityembedding$, scale-adaptive upsampling block, sub-point convolution and upsampling refinement layer, are then added incrementally, as shown in \Table{component}.
All the modules contribute to the reconstruction quality under a fixed Bpp.

\begin{table}[t!]
\centering
\vspace{-5pt}
 \resizebox{.85\linewidth}{!}{
\begin{tabular}{lccc}
\toprule    
Components & CD ($10^{-2}$) $\downarrow$  & PSNR $\uparrow$ & DM $\downarrow$ \\
\midrule
Baseline &  2.61 & 38.82 & 4.17 \\
+$\upsamplenumber$ & 2.29 & 39.64 & 3.23 \\
+$\positionembedding$ & 1.67 & 40.96 & 3.02 \\
+$\densityembedding$ & 1.32 & 41.68 & 2.58 \\
+Adaptive Scale & 0.98 & 42.49 & 2.31 \\
+Sub-point Conv & 0.45 & 43.73 & 2.07 \\
+Refinement & \textbf{0.36} & \textbf{44.03} & \textbf{1.98} \\
\bottomrule
\end{tabular}}
\vspace{-5pt}
\caption{\label{tab:component} The effectiveness of each component in our method. Each row  a component is added on top of the previous row. }
\vspace{-10pt}
\end{table}

\paragraph{Effectiveness of our decoder}
To show that our decoder, consisting of our upsampling block and sub-point convolution, is more effective in leveraging the information provided by the encoder for recovering density, we utilize various point upampling modules from previous works as the decoders to jointly train with our encoder, as shown in \Table{upsampling_block}. 
Our decoder significantly outperforms others on all the reconstruction quality metrics, indicating that our decoder preserves geometry and local density better.

\begin{table}[htbp]
\vspace{-5pt}
\centering
  \resizebox{.85\linewidth}{!}{
\begin{tabular}{lccc}
\toprule
Decoders & CD ($10^{-2}$) $\downarrow$  & PSNR $\uparrow$ & DM $\downarrow$ \\
\midrule
Yu \etal \cite{yu2018pu} & 1.25 & 41.51 & 2.60 \\
Wang \etal \cite{yifan2019patch} & 1.03 & 42.54 & 2.46 \\
Li \etal \cite{li2019pu} & 0.98 & 42.57 & 2.45 \\
Li \etal \cite{li2021point} & 0.90 & 42.83 & 2.32 \\
Qian \etal \cite{qian2021pu} & 0.81 & 43.06 & 2.25 \\
\midrule
Ours & \textbf{0.36} & \textbf{44.03} & \textbf{1.98} \\
\bottomrule
\end{tabular}}
\vspace{-5pt}
 \caption{\label{tab:upsampling_block} The effectiveness of our decoder. 
 In each row, we replace our decoder with the decoder from another work. }
\vspace{-10pt}
\end{table}

\begin{figure}[htbp]
\flushleft
\vspace{-5pt}
\includegraphics[trim={0cm 0.4cm 0cm 0cm}, clip, width=\columnwidth]{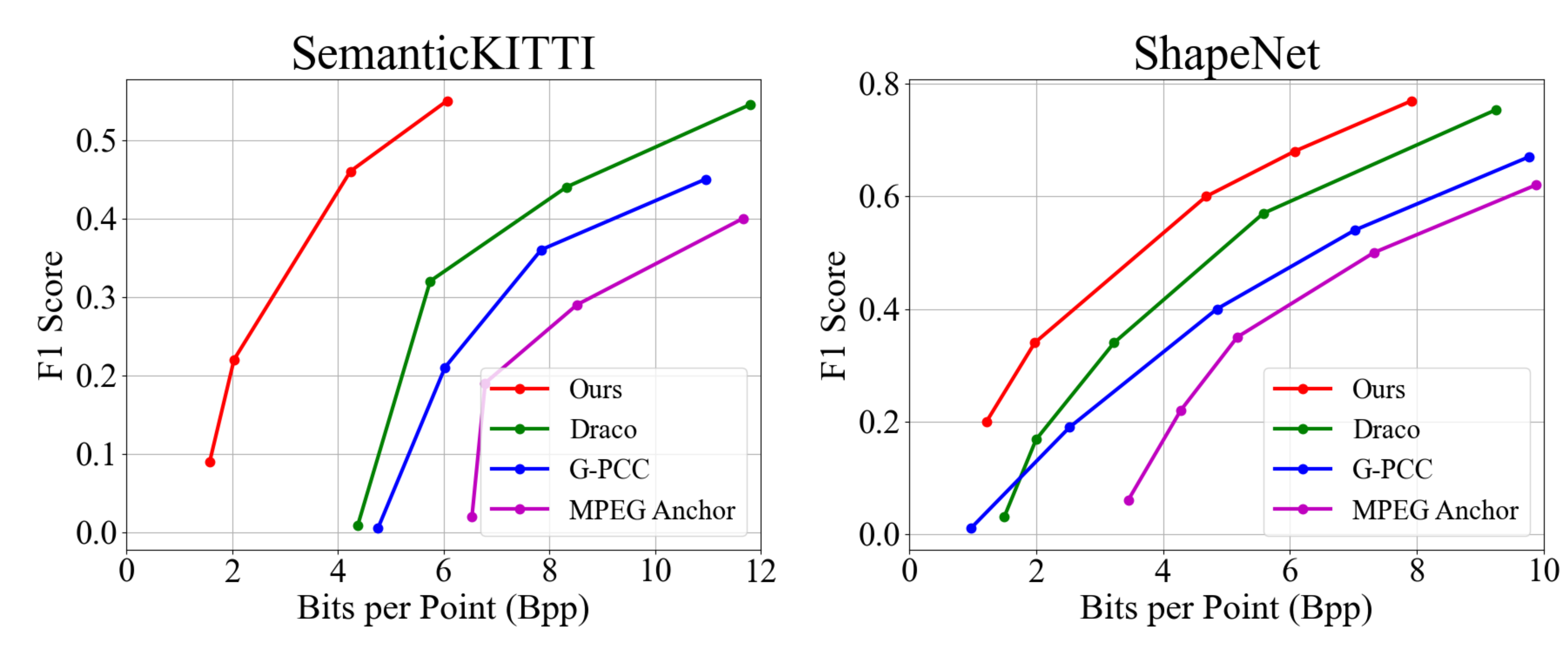}
\caption{Quantitative normal compression results. Left: SemanticKITTI; right: ShapeNet. Our method consistently performs better than Draco \cite{galligan2018google}, G-PCC \cite{graziosi2020overview} and MPEG Anchor \cite{mekuria2016design} across the bitrate spectrum. }
\label{fig:normal_quantity}
\vspace{-10pt}
\end{figure}

\subsection{Normal Compression}
Besides positions, we also evaluate the capability of compressing attributes, using normals as an example.
The normals are concatenated with the point locations and fed into our model. The decompressed locations and normals are then compared with the inputs by per-block F1 score~\cite{biswas2020muscle}. 

As modifying learning based approaches such as PCGC \cite{wang2021lossy}
and Depoco \cite{wiesmann2021deep} to have attribute compression is non-trivial, we only compare to Draco \cite{galligan2018google}, G-PCC \cite{graziosi2020overview} and MPEG Anchor \cite{mekuria2016design}, as shown in \Fig{normal_quantity}.
Our method consistently outperforms others, especially by a large margin on the SemanticKITTI dataset.

\begin{figure}[t!]
\centering
\vspace{-10pt}
\includegraphics[trim={0cm 0.4cm 0cm 0cm}, clip,width=\linewidth]{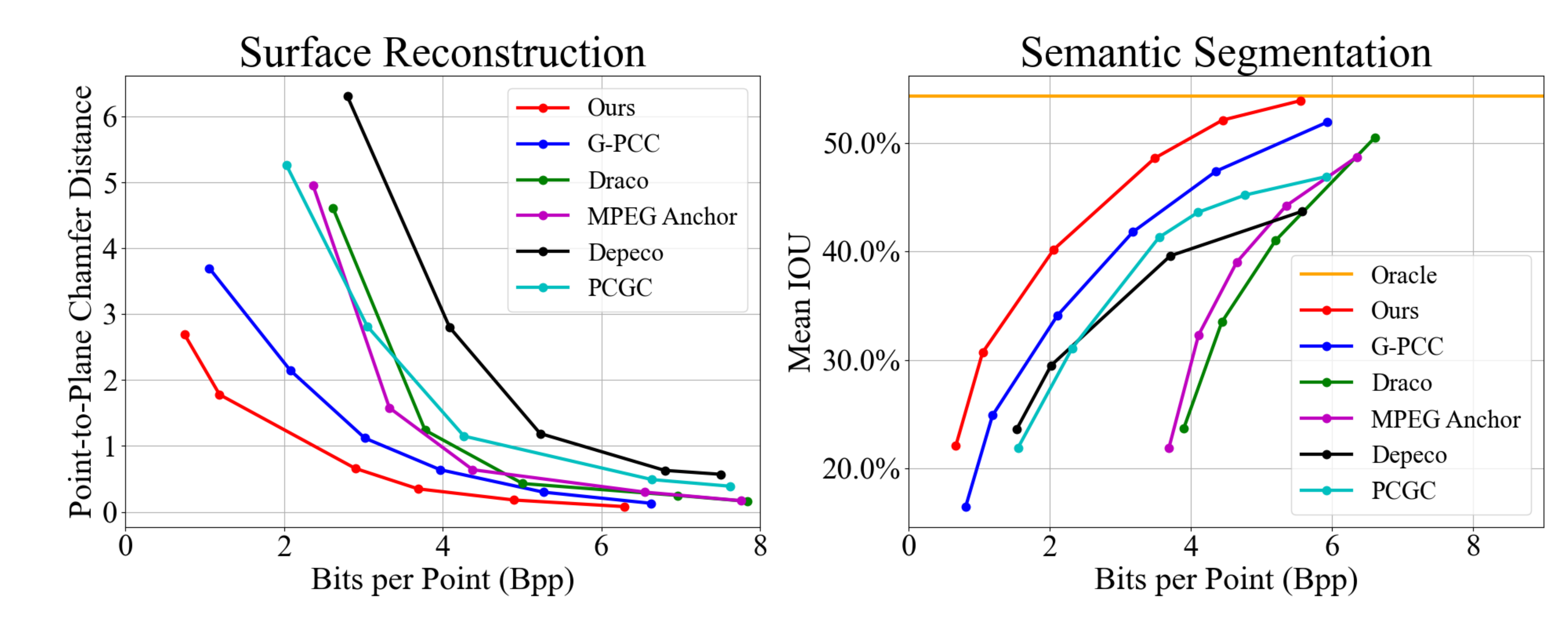}
\caption{Quantitative results of downstream tasks. Left: surface reconstruction on RenderPeople; right: semantic segmentation on SemanticKITTI. }
\label{fig:downstream_task}
\vspace{-10pt}
\end{figure}

\subsection{Impact on Downstream Tasks}
Point cloud compression, as an upstream task, should not affect the performance of downstream applications much. 
In this section, we compare the impact of different compression algorithms on two downstream tasks: surface reconstruction and semantic segmentation. Since some methods do not support attribute compression, all methods only compress the positions for fair comparison.

In the surface reconstruction experiments, Poisson reconstruction \cite{kazhdan2013screened} is run on the full decompressed point clouds. Reconstructed meshes are then compared with the ground truth with the symmetric point-to-plane Chamfer Distance \cite{tang2020deep}. 
For semantic segmentation, we train PolarNet \cite{zhang2020polarnet} on raw point clouds from SemanticKITTI training set, and test on the full decompressed point clouds. The mean intersection-over-union (IOU) is used as metric, following~\cite{huang2020octsqueeze}.
As shown in \Fig{downstream_task}, our method consistently yields the best rate-distortion trade-off, which reiterates the importance of recovering local density. Please refer to the supplementary section for qualitative comparisons.

\section{Conclusion}

We introduce a novel deep point cloud compression framework that can preserve local density. 
Not only does it yield the best rate-distortion trade-off against prior arts, it also recovers local density more accurately under our density metric. Qualitative results show that our algorithm can mitigate the two main density issues of other methods: uniformly distributed and clustered points.
Complexity wise our method is only second to Depoco while with much better accuracy.

\paragraph{Acknowledgments}
This work was supported in part by NSFC under Grant (No. 62076067), SMSTM Project (2021SHZDZX0103), and Shanghai Research and Innovation Functional Program (17DZ2260900). Danhang Tang, Yinda Zhang and Yanwei Fu are the corresponding authours.

\clearpage
{\small
\newpage
\bibliographystyle{ieee_fullname}
\bibliography{main}
}

\clearpage
\noindent \textbf{\Large Supplementary Material}
\setcounter{section}{0}
\vspace{+10pt}

In the supplementary material, we provide additional implementation details, ablation studies and qualitative results.
Limitation and potential ethic concerns are also discussed.

\section{Implementation Details}
Additional details about hyperparameter settings, detailed network architecture, and diagrams of baselines used in ablation study are elucidate in this section. We also formulate the reconstruction metrics used in experiments.

\subsection{Hyperparameters}
In the experiments, we choose the number of stages $\totalstage=3$ and the downsampling factor $\downsamplefactor_\stage \in \{1/2, 1/3, 1/4\}$.
We set dimension $\dimension=8$ for all three embeddings extracted by the encoder, and maximum upsampling factor $\maxupsamplenumber=8$ in the decoder. 
For distortion loss, we set the weight of density loss $\weightdensityloss=1e-4$ and cardinality loss $\weightnumpointsloss=5e-7$.
Where in density loss, the coefficient $\weightpointoffsetloss=50$. 
And the weight $\mu$ for density metric is setted as $1e-4$ and $1e-5$ for ShapeNet and SemanticKITTI respectively.
For normal compression, we additionally add a L2 loss between the reconstructed normals and ground truth, and its weight is $1e-2$.
To obtain the rate-distortion trade-off curves, we vary the coefficient of rate loss $\rateweight$ and downsampling factor $\downsamplefactor_\stage$.
Moreover, in the adaptive scale upsampling block, we use an icosahedron to sample uniformly on a unit sphere to get $M=43$ candidate directions, which include 42 vertices of the icosahedron and 1 origin, following \cite{xiao2009image}.

Our model is implemented with PyTorch~\cite{paszke2019pytorch} and CompressAI \cite{begaint2020compressai}, trained on the NVIDIA TITAN X GPU for 50 epochs. 
We use the Adam optimizer~\cite{kingma2014adam} with a initial learning rate of 1e-3 and a decay factor of 0.5 every 15 epochs.

\subsection{Detailed Network Architecture}
\begin{figure*}[t!]
\centering
\includegraphics[width=\textwidth]{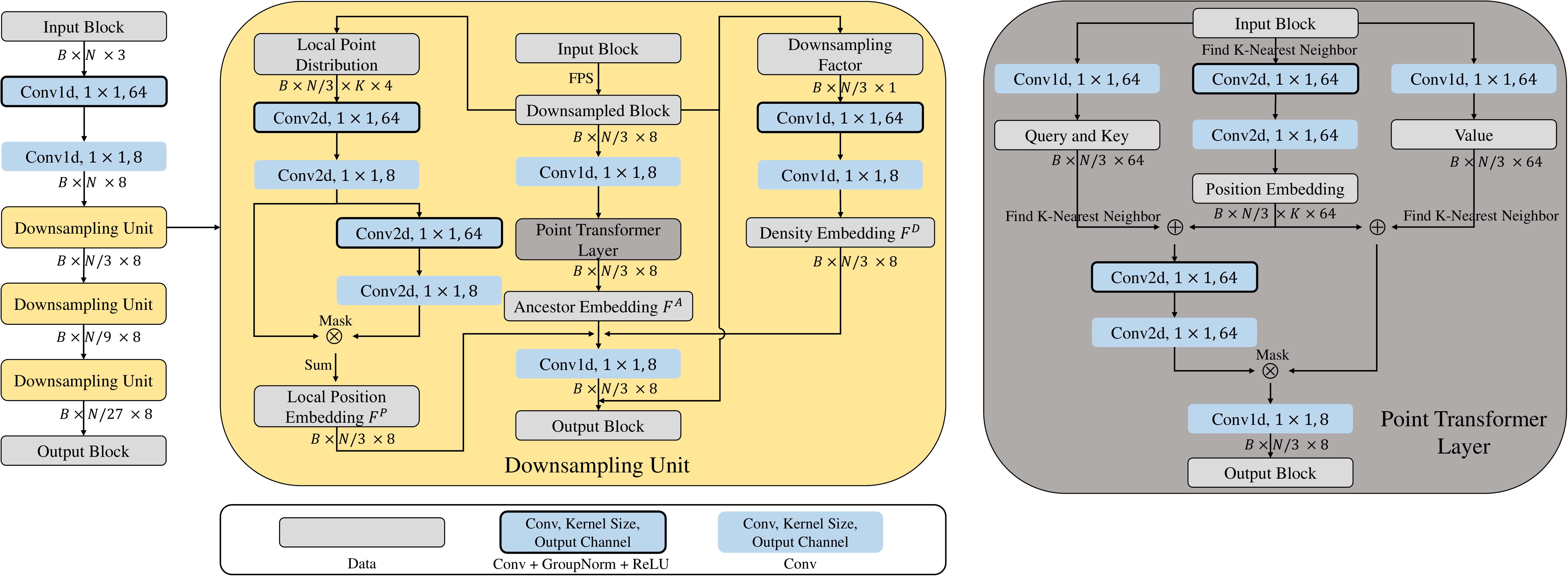}
\caption{The detailed architecture of our encoder. The text below each box indicates the feature dimension: batchsize $\times$ points number $\times$ (k-nearest neighbor) $\times$ channel. And the stride of all convolution layers is fixed to 1. }
\label{fig:encoder}
\end{figure*}

\begin{figure*}[t!]
\centering
\includegraphics[width=\textwidth]{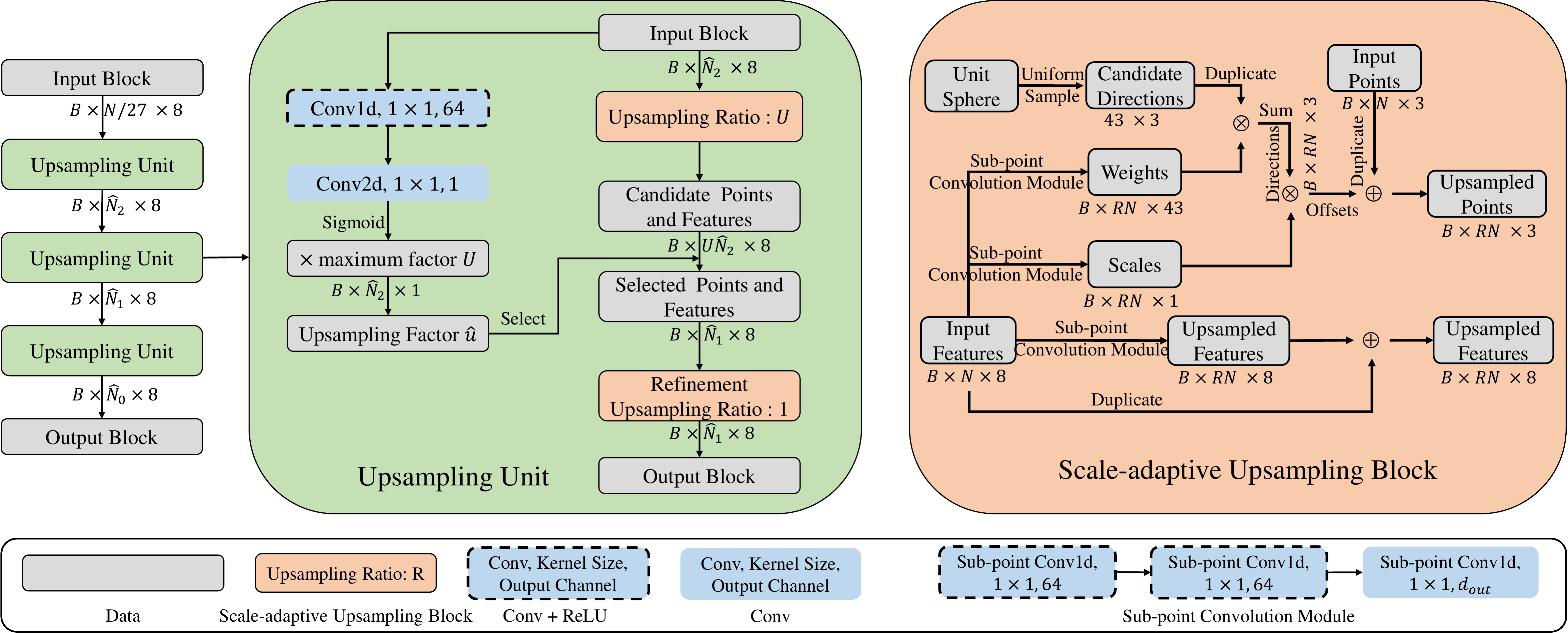}
\caption{The detailed architecture of our decoder. The text below each box indicates the feature dimension: batchsize $\times$ points number $\times$ (k-nearest neighbor) $\times$ channel,
where $\hat{N}_2 \approx N/9, \hat{N}_1 \approx N/3, \hat{N}_0 \approx N$.
And the stride of all convolution layers is fixed to 1.}
\label{fig:decoder}
\end{figure*}

The detailed architectures of our encoder and decoder are shown in \Fig{encoder} and \Fig{decoder} separately.
At stage $\stage$ of the encoder, for each point $\point \in \pointcloud_{\stage+1}$, we first extract local position embedding $\positionembedding$ and density embedding $\densityembedding$ to capture the local geometry and density information of current stage. 
And ancestor embedding $\ancestorembedding$ is also utilized to aggregate features from previous stages by applying point transformer layer \cite{zhao2021point}, based on the collapsed points set $\collapsedpointsset(\point)$.
However, the cardinality of each point's collapsed set may be different.
To achieve parallel process, we first find the k-nearest neighbor $\mathcal{K}(\point)$ in $\pointcloud_{\stage}$ for each downsampled point $\point$, and then apply a mask when using attention for feature aggregation.
Specifically, the mask is defined as below:

\begin{equation}
    w_k=
    \begin{cases}
    MLPs(\feature(\point_k)), & \textrm{if}~ \point_k \in \mathcal{K}(\point)~ \textrm{and}~ \point_k \in \collapsedpointsset(p)  \\[2ex]
    0, & \textrm{else}
    \end{cases}
    \label{eq:mask}
\end{equation}
where $w_k$ and $\feature(\point_k)$ are the weight and feature of $\point_k$.
We set $k=16$, which is much larger than $1/\downsamplefactor_\stage$, so the collapsed set $\collapsedpointsset(\point)$ is guaranteed to be the subset of $\mathcal{K}(\point)$.

And in the decoder, we apply sub-point convolution to construct the scale-adaptive upsampling block, which promotes the recovering of local geometry patterns and density.

\subsection{Reconstruction Metrics}
We use the symmetric point-to-point Chamfer Distance and point-to-plane PSNR to evaluate the geometry accuracy of reconstructed point clouds.
And now we list their mathematical formulas.

Given ground truth $\pointcloud_{\stage}$ and reconstructed point cloud $\reconstructedpointcloud_{\stage}$, the calculation of symmetric point-to-point Chamfer Distance is as follow:

{\footnotesize
\begin{equation}
\begin{aligned}
    CD(\pointcloud_{\stage}, \reconstructedpointcloud_{\stage}) =\frac{1}{|\pointcloud_{\stage}|} \sum_{p \in \pointcloud_{\stage}} \min _{\hat{p} \in \reconstructedpointcloud_{\stage}}\|p-\hat{p}\|_{2}^{2}+
    \frac{1}{|\reconstructedpointcloud_{\stage}|} \sum_{\hat{p} \in \reconstructedpointcloud_{\stage}} \min _{p \in \pointcloud_{\stage}}\|\hat{p}-p\|_{2}^{2}
\end{aligned}
\end{equation}
}

Following \cite{biswas2020muscle}, we calculate the symmetric point-to-plane PSNR as:

{\footnotesize
\begin{equation}
\begin{aligned}
     PSNR(\pointcloud_{\stage}, \reconstructedpointcloud_{\stage}) =10 \log _{10} {\frac{3 \sigma^2}{\max \{ MSE(\pointcloud_{\stage}, \reconstructedpointcloud_{\stage}), MSE(\reconstructedpointcloud_{\stage}, \pointcloud_{\stage}) \}}}
\end{aligned}
\end{equation}}where $\sigma$ is the peak constant value, 
represented by the maximum nearest neighbor distance in the whole dataset \cite{biswas2020muscle}.
$MSE(\pointcloud_{\stage}, \reconstructedpointcloud_{\stage})=\frac{1}{|\pointcloud_{\stage}|} \sum_{\point \in \pointcloud_{\stage}}((\point-\reconstructedpoint) \cdot \hat{n})^{2}$,
where $\reconstructedpoint$ is $\point$'s nearest neighbor in $\reconstructedpointcloud_{\stage}$, and $\hat{n}$ is the normal of $\reconstructedpoint$.
For each $\point \in \pointcloud_{\stage}$, we estimate its normal using \cite{zhou2018open3d}.
And for each $\reconstructedpoint \in \reconstructedpointcloud_{\stage}$, we use the normal of its nearest neighbor in  $\pointcloud_{\stage}$ as $\hat{n}$.

While Chamfer Distance and PSNR can only be used for position compression, we apply F1 score to measure the quality of both reconstructed locations and normals during normal compression, based on \cite{biswas2020muscle}.

{\small
\begin{equation}
    {F}_{1}(\pointcloud_\stage, \reconstructedpointcloud_\stage)=\frac{2 TP}{2TP+FP+FN}
\end{equation}
}where $TP$ (true positives) represent those reconstructed points $(\reconstructedpoint, \hat{n}) \in \reconstructedpointcloud_{\stage}$ which have a corresponding ground truth point $(\point, n) \in \pointcloud_{\stage}$ that satisfies $||\point-\reconstructedpoint||_{2} \leq \tau_{p}$ and $||{n}-\hat{n}||_{2} \leq \tau_{n}$;
$FP$ (false positives) indicate the rest reconstructed points; and $FN$ (false negatives) are those ground truth points which do not have a corresponding $TP$.
For SemanticKITTI, we set $\tau_{p}=0.5, \tau_{n}=0.5$; and for ShapeNet, we set $\tau_{p}=0.05, \tau_{n}=0.2$.

\subsection{Baselines in Ablation Study}
\begin{figure*}[t!]
\centering
\includegraphics[width=0.61\textwidth]{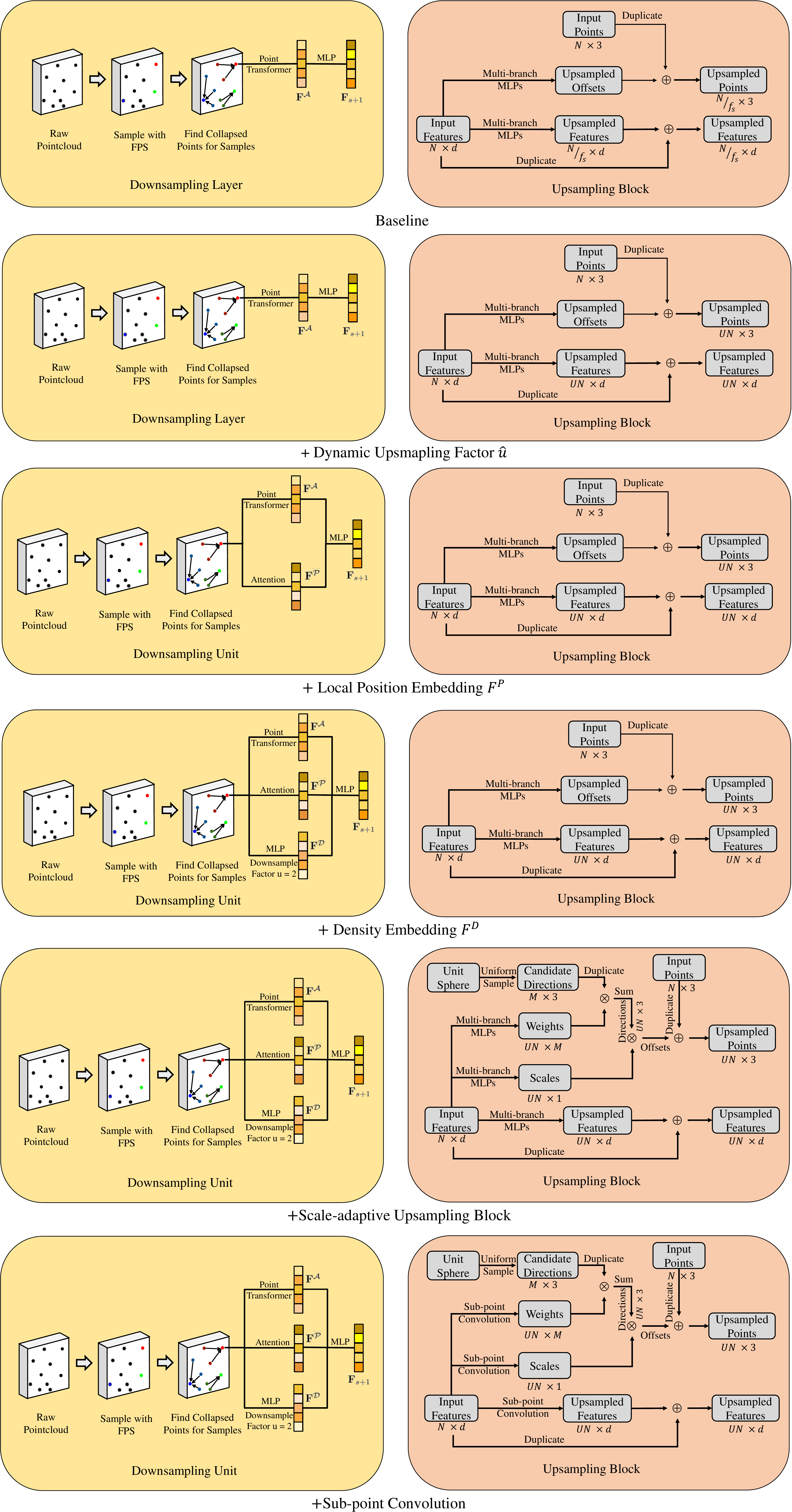}
\caption{The detailed structures in ablation study (Table 2 of the main paper).
Left: alternatives for the downsampling block in the encoder; right: alternatives for the scale-adaptive upsampling block in the decoder.
For dynamic upsampling factor, we first generate $\maxupsamplenumber$ items and then select the first $\upsamplenumber$ points and features.
While all these baselines do not have the refinement layer in the decoder, our full model adds it based on the ``+Sub-point Convolution" model. 
}
\label{fig:baseline}
\end{figure*}

In the Table 2 of main paper, we validate the effectiveness of each component in our method.
To achieve so, we first build a baseline model, which is composed of a point transformer encoder \cite{zhao2021point}, entropy encoder and multi-branch MLPs decoder \cite{yu2018pu}.
And we utilize a fixed upsampling factor $1/\downsamplefactor_\stage$ for this baseline.
Then we add the following components incrementally: dynamic upsampling factor $\upsamplenumber$, local position embedding $\positionembedding$, density embedding $\densityembedding$, scale-adaptive upsampling block, sub-point convolution and upsampling refinement layer.
Here we draw the detailed structures of each model, as shown in \Fig{baseline}.
Note that for all these models, we adopt the same pipeline, and only enable our contributing component once a time.

\section{Additional Ablation Studies}
In this section, we conduct some more ablation experiments to validate our choices of downsampling methods and loss functions.
And all these experiments are conducted on SemanticKITTI with fixed bpp 2.1, the same as the main paper.

\subsection{Downsampling Methods} 
At stage $\stage$ of the encoder, we use FPS to get the downsampled point cloud $\pointcloud_{\stage+1}$, which expected to have a good coverage of the input $\pointcloud_{\stage}$.
Besides FPS, there are also two common downsampling methods: random downsampling (RD) and grid downsampling (GD) \cite{wiesmann2021deep}.
And we replace FPS with these two downsampling methods in turn, as shown in \Table{downsampling}.
It is obvious that FPS can significantly improve the accuracy of reconstruction because it has better coverage, both in terms of geometry and local density.

\begin{table}[htbp]
\centering
\resizebox{.85\linewidth}{!}{
 \begin{tabular}{lccc}
  \toprule
\begin{tabular}[c]{@{}c@{}} Downsampling \\ Methods \end{tabular} & CD ($10^{-2}$) $\downarrow$  & PSNR $\uparrow$ & DM $\downarrow$ \\
  \midrule
RD &  1.29          & 41.17 & 3.08 \\
GD &  0.62          & 42.86 & 2.31 \\
FPS & \textbf{0.36}  & \textbf{44.03} & \textbf{1.98} \\
  \bottomrule
 \end{tabular}}
  \caption{\label{tab:downsampling}The effectiveness of different downsampling methods. 
 It is clear that FPS delivers the best performance.}
 \vspace{-10pt}
\end{table}

\subsection{Loss Functions}
In our framework, we adopt the standard rate-distortion loss function for training.
And the symmetric point-to-point Chamfer Distance $\chamferloss$ is used as the distortion loss $\distortionloss$, while the estimated bits number is used as the rate loss $\rateloss$.
In addition to these two loss functions, we also extend the distortion loss by designing the density loss $\densityloss$ and cardinality loss $\numpointsloss$ to facilitate the recovery of local density.
For validating the new loss functions $\densityloss$ and $\numpointsloss$, we remove them degressively, as shown in \Table{loss}.

\begin{table}[htbp]
\centering
\resizebox{.9\linewidth}{!}{
 \begin{tabular}{lccc}
  \toprule
  Loss Functions & CD ($10^{-2}$) $\downarrow$  & PSNR $\uparrow$ & DM $\downarrow$ \\
\midrule
    Full Loss Functions & \textbf{0.36} & \textbf{44.03} & \textbf{1.98} \\
    -$\numpointsloss$ & 0.47 & 43.62 & 2.15 \\
    -$\densityloss$ & 1.45 & 40.74 & 3.59 \\
  \bottomrule
 \end{tabular}}
  \caption{\label{tab:loss}The effectiveness of loss functions. Each row  a loss function is romoved based on the top of previous row.}
  \vspace{-5pt}
\end{table}

As cardinality loss $\numpointsloss$ is removed, all metrics drop slightly.
However, once the constrain of local density is absent, the reconstruction quality will drop sharply, indicating the effectiveness of our designed density loss.

\section{Additional Qualitative Results}
\begin{figure*}[t!]
\centering
\includegraphics[width=0.95\textwidth]{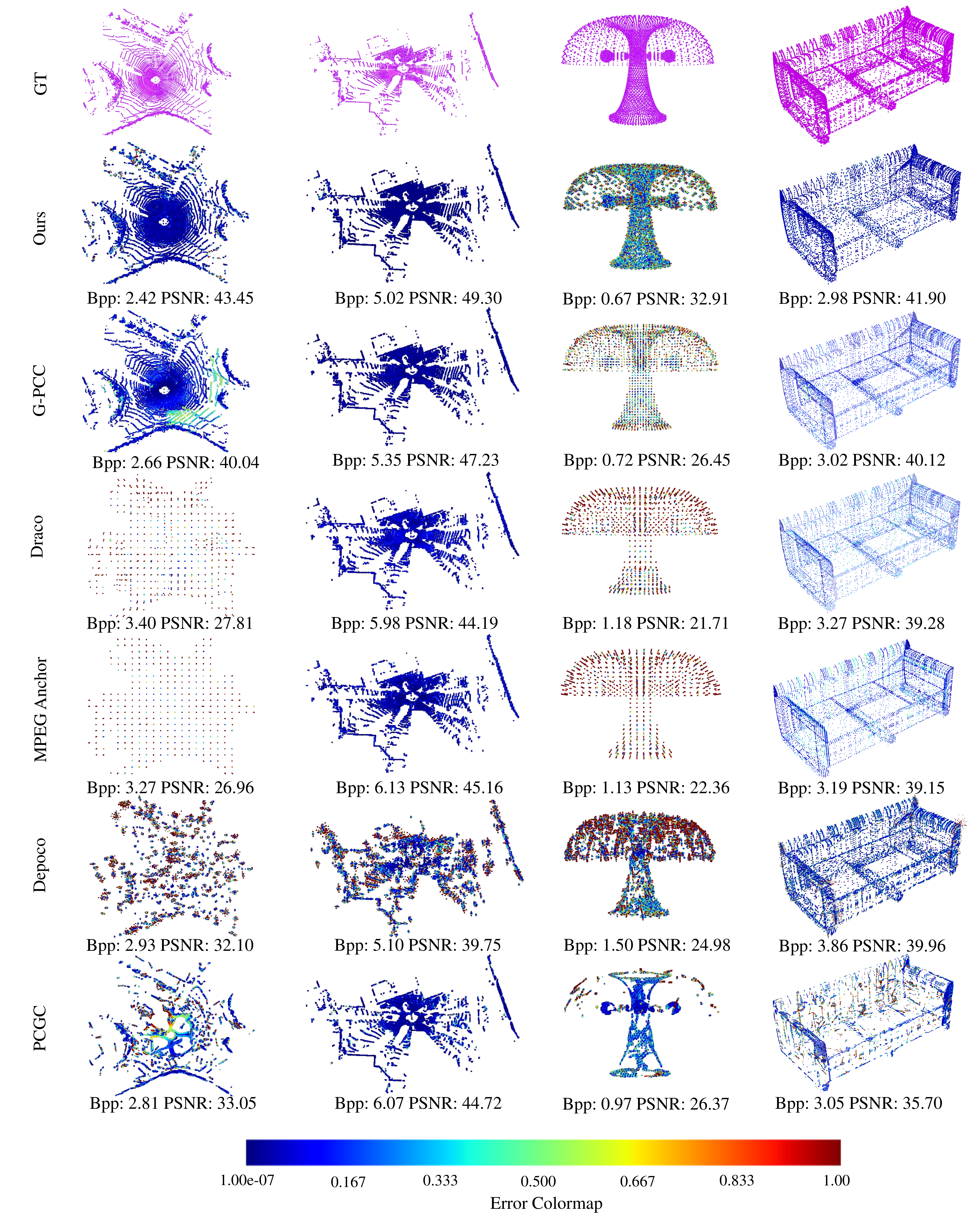}
\caption{More qualitative results on SemanticKITTI (the first two coloums) and ShapeNet (the last two coloums).
From top to bottom: Ground Truth, Ours, G-PCC~\cite{graziosi2020overview}, Draco~\cite{galligan2018google}, MPEG Anchor \cite{mekuria2016design}, Depeco~\cite{wiesmann2021deep} and PCGC~\cite{wang2021lossy}.
We utilize the distance between each point in decompressed point clouds and its nearest neighbor in ground truth as the error.
And the Bpp and PSNR metrics are averaged by each block of the full point clouds.}
\label{fig:more_quality}
\end{figure*}

\begin{figure*}[t!]
\centering
\includegraphics[width=\textwidth]{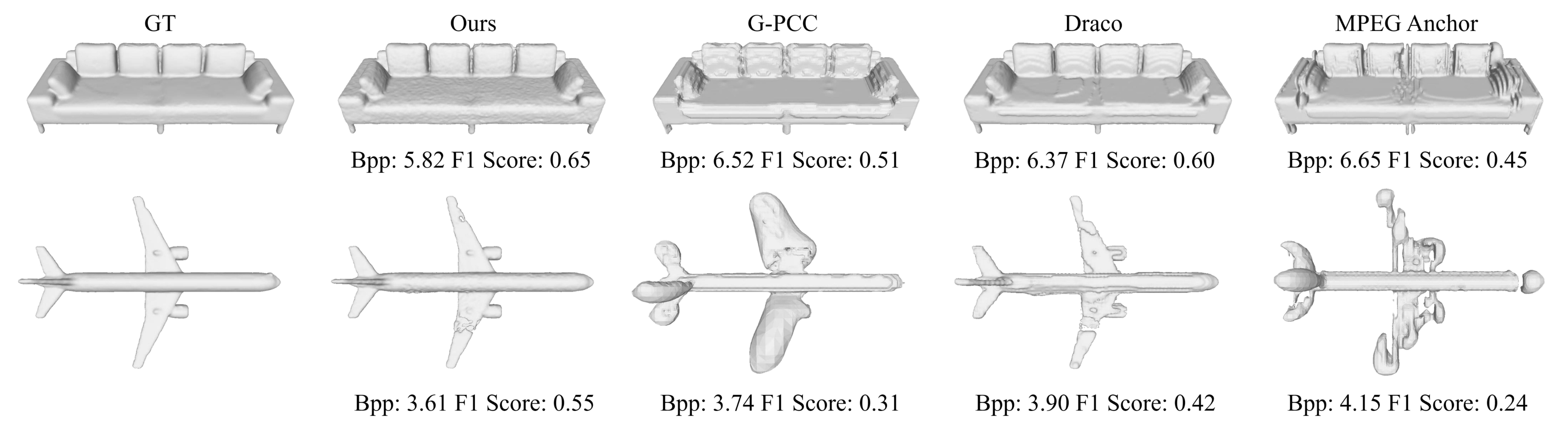}
\caption{Qualitative results of normal compression. We apply Poisson reconstruction \cite{kazhdan2013screened} to generate the mesh based on decompressed points and normals.
And the Bpp and PSNR metrics are averaged by each block of the full point clouds.
}
\label{fig:normal_quality}
\end{figure*}

\begin{figure*}[t!]
\centering
\includegraphics[width=\textwidth]{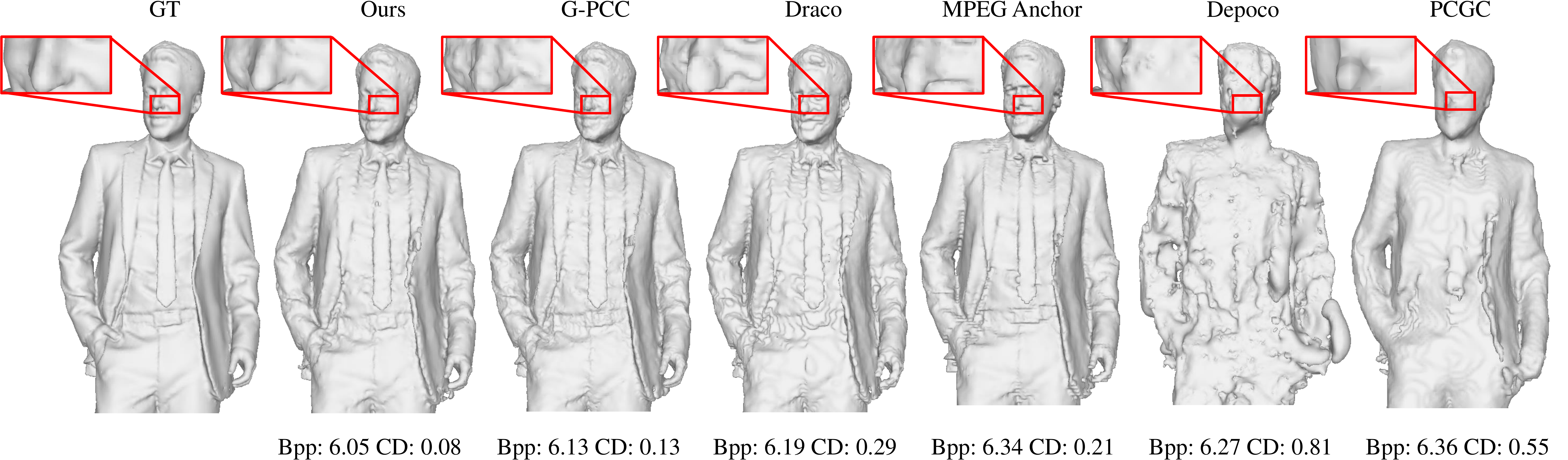}
\caption{Qualitative results on the surface reconstruction downstream task, where CD represents the symmetric point-to-plane Chamfer Distance \cite{tang2020deep} on each full model.
It is clear that the mesh reconstructed from our decompressed point cloud contains better surface details
and more accurate geometry than others, especially on the face.
}
\label{fig:surface_reconstruction}
\end{figure*}

\begin{figure*}[t!]
\centering
\includegraphics[width=\textwidth]{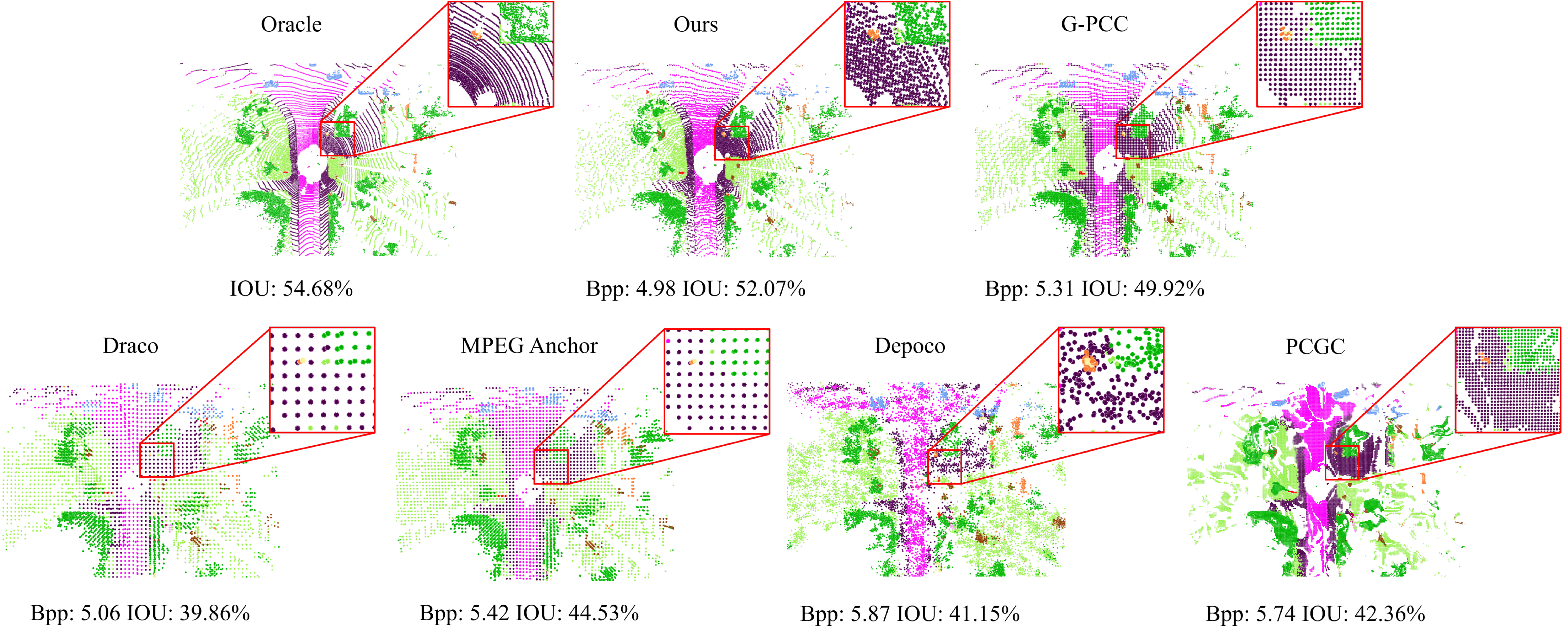}
\caption{Qualitative results on the semantic segmentation downstream task, where IOU denotes the intersection-over-union metric on each scan.
It is shown that preserving local density not only recovers more accurate geometry, but also benefits downstream task.
}
\label{fig:segmentation}
\end{figure*}

In this section, we show more qualitative results on position compression, normal compression and downstream tasks, which clearly indicate that our density-preserving compression approach achieves the best performance.
Specifically, in \Fig{more_quality}, we show more qualitative position compression results on SemanticKITTI and ShapeNet.
In \Fig{normal_quality}, we visualize the normal compression results by employing Poisson reconstruction \cite{kazhdan2013screened} on decompressed points and normals.
In \Fig{surface_reconstruction} and \Fig{segmentation}, we display the qualitative results of two downstream tasks: surface reconstruction and semantic segmentation.

\section{Limitation Discussion}
\begin{figure*}[t!]
\centering
\includegraphics[width=\textwidth]{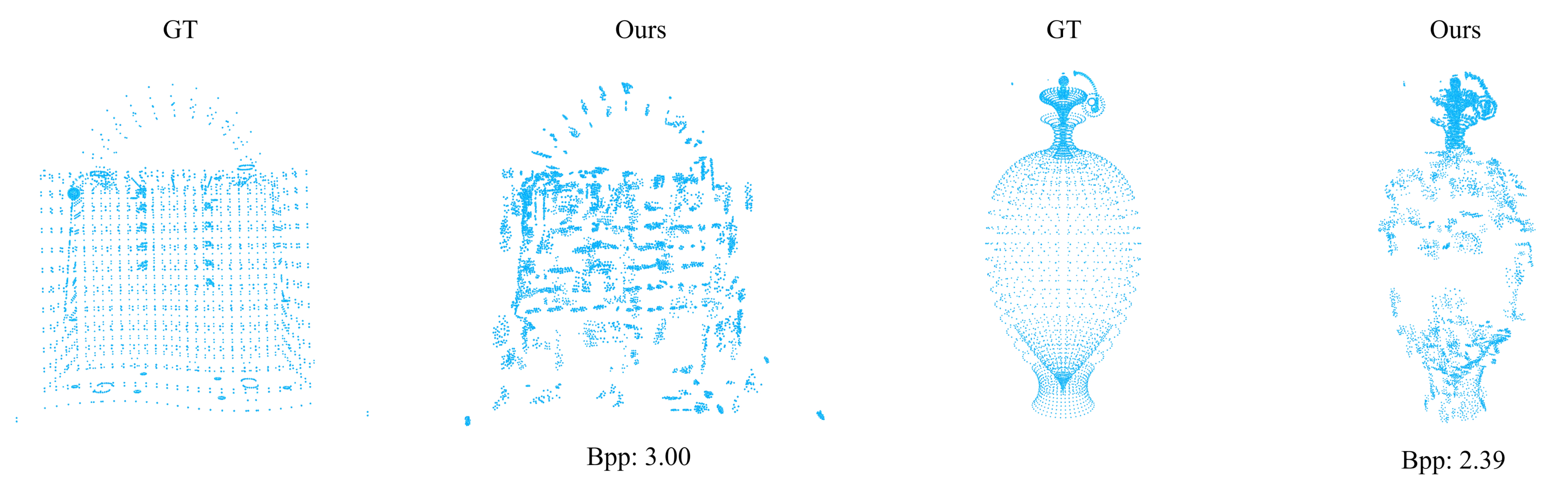}
\caption{Failure cases on extremely sparse point clouds. 
}
\label{fig:failure}
\end{figure*}

Although our density-preserving deep point cloud compression framework is effective, it also has some limitations. 
For example: 1) The maximum upsampling factor $\maxupsamplenumber$ is predefined before decoding, thus the actual upsampling factor $\upsamplenumber$ is expected to be less than or equal to $\maxupsamplenumber$.
However, the assumption may be broken in some cases, especially when the local area is extremely dense, 
then our method may not be able to recover the local density precisely.
2) As we divide the point clouds into small blocks, each block may contain various number of points, so they are not easy to perfectly parallelized.
3) Other hyperparameters such as the weight of loss, the dimension of embedding, etc may be adaptively adjusted on different datasets.
Moreover, we show some failure cases on extremely sparse point clouds in \Fig{failure}.
As we assume that there exists some data redundancy in the local areas of point clouds, so we can compress it while achieving tolerable distortion.
However, this assumption may not hold when the point cloud is very sparse, and even the downsampled point cloud cannot describe the underlying geometry any more, hence is hard for reconstruction.

At last, we discuss the possible ethical issues. In general, since our point cloud compression algorithm is agnostic to the contents of point clouds, the responsibility of handling ethical issues belongs to the point cloud creator. 
That being said, as compressed point clouds may be intercepted by hackers during network transmission, which may result in data leakage, common encryption algorithms can be applied on the bottleneck point clouds and features to protect user privacy.

\clearpage

\end{document}